\documentclass[a4paper, 12pt]{article}

\usepackage[utf8]{inputenc} % allow utf-8 input
\usepackage[T1]{fontenc}    % use 8-bit T1 fonts
\usepackage{hyperref}       % hyperlinks
\usepackage{url}            % simple URL typesetting
\usepackage{booktabs}       % professional-quality tables
\usepackage{amsfonts}       % blackboard math symbols
\usepackage{nicefrac}       % compact symbols for 1/2, etc.
\usepackage{microtype}      % microtypography
\usepackage{xcolor}         % colors

\usepackage{amsmath, amssymb, amsfonts}  % Math stuff
\usepackage{cases}
\usepackage{fullpage}

\usepackage{graphicx}    
\usepackage{bm}
\usepackage{bbm}
\usepackage{dsfont}
\usepackage{enumerate}

\usepackage{ifpdf}

\usepackage[mathscr]{eucal}
\usepackage{subfigure}
\usepackage{theorem}

\usepackage{algorithm}
\usepackage{algpseudocode}

\newtheorem{theorem}{Theorem}

\newtheorem{proposition}{Proposition}
\newtheorem{lemma}{Lemma}
\newtheorem{corollary}{Corollary}

{\theorembodyfont{\upshape}

}

\def\qed{\hfill \vrule height 5pt width 5pt depth 0pt \medskip}

\def\diag{{\rm diag}}
\def\sign{{\rm sgn}}

\newcommand{\beq}{\begin{equation}}
\newcommand{\eeq}{\end{equation}}
\newcommand{\beqa}{\begin{eqnarray}}
\newcommand{\eeqa}{\end{eqnarray}}
\newcommand{\beqan}{\begin{eqnarray*}}
\newcommand{\eeqan}{\end{eqnarray*}}

\newcommand{\pde}[2]{ \frac{\partial #1}{\partial #2} }

\newcommand{\bite}{\begin{itemize}}
\newcommand{\eite}{\end{itemize}}
\newcommand{\benu}{\begin{enumerate}}
\newcommand{\eenu}{\end{enumerate}}

\begin{document}

\title{\LARGE \bf Computing frustration and near-monotonicity in deep neural networks}

\date{October 6, 2025}

\author{Joel Wendin\thanks{Division of Automatic Control, Dept. of Electrical Engineering, Link\"oping University, SE-58183, Link\"oping, Sweden.}, Erik G. Larsson\thanks{Division of Communication Systems, Dept. of Electrical Engineering, Link\"oping University, SE-58183, Link\"oping, Sweden.}, and Claudio Altafini\footnotemark[1] \thanks{Corresponding author (email: {\tt\small claudio.altafini@liu.se})}}

\maketitle

\begin{abstract}

For the signed graph associated to a deep neural network, one can compute the frustration level, i.e., test how close or distant the graph is to structural balance. 
For all the pretrained deep convolutional neural networks we consider, we find that the frustration is always less than expected from null models.
From a statistical physics point of view, and in particular in reference to an Ising spin glass model, the reduced frustration indicates that the amount of disorder encoded in the network is less than in the null models. 
From a functional point of view, low frustration (i.e., proximity to structural balance) means that the function representing the network behaves near-monotonically, i.e., more similarly to a monotone function than in the null models. 
Evidence of near-monotonic behavior along the partial order determined by frustration is observed for all networks we consider.
This confirms that the class of deep convolutional neural networks tends to have a more ordered behavior than expected from null models, and suggests a novel form of implicit regularization.
\end{abstract}

\section*{Introduction}

Even though current deep neural networks (DNN) can achieve human-like performances in fields such as image classification, speech recognition and language generation, the reasons and properties underpinning these successes remain largely unclear. 
Factors such as overparametrization and the possibility of training on massive datasets are certainly playing a role, but alone do not provide a justification of the generalization power of DNNs, i.e., of how even elementary training algorithms like gradient descent tend to produce models that generalize well to new data without signs of overfitting, even in absence of any explicit regularization. 

Very little is known in general of the organization principles that enable a DNN to function so efficiently. 
In simpler settings such as deep linear networks \cite{achour2021loss,arora2018optimization,chitour2018geometric,saxe2013exact,yun2017global}, it is known for instance that lack of convexity in the loss function does not imply existence of local minima in the loss landscape \cite{kawaguchi2016deep}, which guarantees that a reasonably trained linear network typically achieves performances that are optimal within the class of linear models.
Nevertheless, even in this simplified setting, it is unclear if there is any principle governing the ``inner functioning'' of the trained network. 
In matrix factorization problems, for instance, the models seem to obey a ``greedy low-rank'' principle, at least when they are initialized near the origin \cite{cohen2024lecture,li2020towards}. 
In the nonlinear case, the theoretical understanding is even more limited. 
For classification problems with separable data, and binary classification on non-separable data, it appears that a trained model obeys to a maximum-margin principle \cite{ravi2024implicit,soudry2018implicit,ji2019implicit}, but how to extend this concept to more general settings is unclear.
The paper \cite{choromanska2015loss} discusses the loss landscape of general deep neural networks with ReLU activations, using results from random matrix theory applied to spherical spin glasses. In particular, it shows that, under certain assumptions, the low-value critical points of the DNN are located into a narrow band lower bounded by the global minimum, and that the number of local minima outside this band decreases exponentially with the size of the networks.
As shown first in \cite{zhang2021understanding}, current large DNNs can achieve zero training error also on randomized labels, expressing the inadequacy of a criterion based on the loss function as a measure of complexity of a model.
This issue is investigated further in e.g. \cite{neyshabur2017exploring,vardi2023implicit}. No satisfactory complexity measure has so far been found for DNNs.

This paper provides a novel ingredient which can shed some light on the internal organization of a trained DNN. 
To do so, we combine notions from Statistical Physics of complex networks, namely the idea of frustration and structural balance from Ising spin glass theory, with the idea of monotonicity of a function, here the DNN map.
The theory of spin glasses is one of the classical theories used to model complex disordered systems in Statistical Physics \cite{mezard1987spin,binder1986spin}, and has been invoked repeatedly to deal with different aspects of neural networks, see e.g. \cite{amit1985spin,choromanska2015loss,dotsenko1995introduction,mezard2009information}. 
Here rather than the spherical spin glass model of \cite{choromanska2015loss}, our reference is to the Ising spin glass model, in which the spin variables at the nodes can assume only a binary value ($ \pm 1 $), instead of any value on a sphere   \cite{binder1986spin,mezard1987spin,mezard2009information}. 
Unlike the spherical model, the Ising spin glass model is characterized by the presence of {\em frustration} \cite{toulouse1987theory}, a graphical property that is taken as a paradigm for the disorder encoded in the spin glass. Frustration-free networks are disorder-free, and are characterized by graphs that are {\em structurally balanced}, meaning that all (undirected) cycles have an even number of negative edges \cite{Harary1960}. 
When the graph is a directed acyclic graph (DAG) as in DNNs, the original notion from Ising spin glasses (where edge weights represent ferromagnetic and antiferromangetic bonds and are typically indirect and binary, $ \pm 1 $) can be extended to weighted signed digraphs without introducing ambiguities.
These broader concepts of frustration and structural balance have been used in the last decade beyond spin glass theory, for general complex networks \cite{Easley10Network}, and applied to other disciplines like Biology \cite{iacono2010monotonicity,ma2008proximity}, Social Sciences \cite{facchetti2011computing,Wasserman94Social}, etc. 
In addition, it has been shown that there is a strict relation between the graphical property of structural balance and a property of dynamical systems called monotonicity \cite{smith1995monotone}. The Jacobian of a monotone dynamical system is in fact structurally balanced everywhere in its state space. 
This relation can be extended from dynamical systems to static functions, like the input-output map of a DNN. 
Monotone DNN have been studied thoroughly because they have the property of being interpretable: increasing a certain input feature is guaranteed to increase/decrease a certain output feature \cite{cano2019monotonic,liu2020certified,runje2023constrained,sharma2020testing,sill1997monotonic}. 
Exploiting the Jacobian of the DNN, the extension of monotonicity from an input-output to an internal (hidden) state perspective is rather straightforward, and allows to verify monotonicity through the structural balance of the DAG associated to the DNN.  

Monotone DNN are however very unlikely to appear when training from large-scale real data, just like signed networks of any kind are unlikely to be exactly structurally balanced. 
A signed graph which is not structurally balanced has some amount of frustration. 
In Ising spin glass theory, this is quantified by computing the ground state of a Hamiltonian energy functional \cite{mezard1987spin}, a problem known to be NP hard \cite{barahona1982computational}. 
Efficient heuristics, inspired by this notion and based on gradient descent principles, were presented e.g. in \cite{iacono2010determining}.
When investigating the amount of frustration present in diverse types of networks from fields such as Biology and Social Sciences, it has often been observed that the networks are less frustrated than expected from null models \cite{facchetti2011computing,iacono2010determining,iacono2010monotonicity,ma2008proximity,tripathi2023minimal}, which can be interpreted as near-monotonicity of the networks \cite{sontag2007monotone}. In other words, these networks appear to encode less disorder than expected from randomizations, and to behave more predictably than completely random models. 

The task of this paper is to check if something similar may be occurring in DNNs. In order to do so we consider state-of-the-art DNNs, like the modern deep convolutional neural networks (CNN) used for image recognition. 
In particular, we focus on several pretrained CNNs, such as SqueezeNet, ShuffleNet, Alexnet, GoogLeNet, ResNet, and compute the associated DAGs. 
On the resulting DAGs, we calculate the level of frustration and compare it with several types of null models, obtained by preserving the convolutional structure at the layers but reshuffling the position of the parameters in the filters, or reshuffling them in a more drastic way that destroys also the convolutional pattern.

All the pretrained CNNs we consider have a certain amount of frustration, which is however always less than that of the null models, and the difference is statistically significant.
Consequently, all pretrained CNNs are expected to have a behavior which is closer to monotone than the associated null models. For our CNNs this is indeed the case: the outputs of the CNNs exhibit a marked tendency to respond to input perturbations by aligning themselves with the partial order direction predicted by our ground state calculations.
This near-monotone tendency is statistically significant for all the CNNs we consider.

Overall, this suggests that the internal organization of a trained CNN encodes some order, which induces a near-monotone input-output behavior never observed before. 
We believe it can be considered a novel form of implicit regularization, i.e., of mathematical simplicity not imposed explicitly but emerging from the training process.

\section*{Methods}
\label{sec:background}
In this section we review the concepts of structural balance and monotonicity, and show how to quantify the distance to structural balance/monotonicity through the notion of frustration.

\subsection*{Structural balance}
\label{sec:background-stru-bal-mon}
Consider a signed digraph $ \mathcal{G}(A)$, where $ A\in \mathbb{R}^{n\times n}$ is the associated weighted adjacency matrix, i.e., $ A_{ij} \neq 0$ iff $\mathcal{G}(A)$ has an edge from node $ j $ to node $i$. The edge weight $ A_{ij} \neq 0$ can be either a positive or negative number. 
Given $ \mathcal{G}(A)$, we denote 
$ \mathcal{G}_u(A)$ the associated undirected graph obtained by dropping the arrow on the edges: $ \mathcal{G}_u(A) = \mathcal{G}(A_u) $, where $ A_u = A+A^\top $ is the symmetrized version of $A$. Since we deal only with DAGs, this operation can always be done without creating ambiguity.
The digraph  $ \mathcal{G}(A)$ is said {\em structurally balanced} (or {\em frustration-free}) if in its associated undirected graph $ \mathcal{G}_u(A )$ all cycles are positive, i.e., they have an even number of negative edges \cite{Harary1960}.\footnote{Self-loops in $ \mathcal{G}(A)$ are normally disregarded. Here this detail can be omitted, as there are never self-loops in our DAGs.
}
If $ \mathcal{G}(A)$ is structurally balanced, then there exists a partition of its nodes into two sets $ \mathcal{V}_1 $ and $ \mathcal{V}_2 $ s.t. $ \mathcal{V}_1 \cup  \mathcal{V}_2 = \{ 1, \ldots, n \} $ and $ \mathcal{V}_1 \cap  \mathcal{V}_2 = \emptyset$, for which $ A_{ij} \geq 0 $ if $ i, j \in \mathcal{V}_1 $ or $ i, j \in \mathcal{V}_2 $, and $ A_{ij}\leq 0 $ if $ i \in \mathcal{V}_1 $ and $ j\in \mathcal{V}_2 $, or $ i \in \mathcal{V}_2 $ and $ j\in \mathcal{V}_1 $.
Another equivalent characterization of structural balance is the existence of a diagonal signature matrix  $ S = \diag(\bm{s})$, where $ \bm{s}= [s_1 \, \ldots \, s_n] $ with $ s_i = \pm 1 $ for all $ i=1, \ldots, n$, s.t. $S A S \geq 0$ \cite{facchetti2011computing}. $S$ is sometimes referred to as a gauge transformation matrix in the Statistical Physics literature.  
The sign of the cycles of $ \mathcal{G}_u(A) = \mathcal{G}(A_u)$ is invariant to gauge transformations by $ S = \diag(\bm{s})$. 
In fact, if $ s_i =-1 $ and $ s_j =+1 $ for all $ j\neq i$, then the effect of the gauge transformation is to flip the sign of all entries in the $i$-th row and column of $A_u$, which leaves the sign of the cycles in which the node $i$ is involved (and of all other cycles as well) invariant, since each (simple) cycle involves 2 edges adjacent to $i$.

\subsection*{Monotonicity}
To introduce the notion of monotonicity we use in this paper, we need to consider the idea of partial ordering induced by an orthant of $ \mathbb{R}^n$. 
Consider the gauge transformation matrix $ S $ and denote $ \mathbb{S}  $ the orthant of $ \mathbb{R}^n $ identified by $S$: $ \mathbb{S} = S \mathbb{R}^n = \{ \bm{z} \in \mathbb{R}^n  \text{ s.t. }  s_i z_i \geq 0, \, i=1, \ldots, n \}$. 
$ \mathbb{S}$ determines the following partial ordering in $ \mathbb{R}^n$, denoted ``$ \leq_\mathbb{S}$'': $ \bm{z}_1 \leq_\mathbb{S} \bm{z}_2 $ iff $ \bm{z}_2-\bm{z}_1 \in \mathbb{S} $ $ \forall \, \bm{z}_1, \, \bm{z}_2 \in \mathbb{R}^n$. 

A function $ f\, :  \, \mathbb{R}^n \to \mathbb{R}^n  $ is said {\em monotone with respect to the partial order $ \mathbb{S}$} (for short $ \mathbb{S}$-monotone, or simply {\em monotone} when there is no need to specify the partial order $ \mathbb{S}$) if $ \bm{z}_1 \leq_\mathbb{S} \bm{z}_2 $ implies $ f(\bm{z}_1) \leq_\mathbb{S} f(\bm{z}_2) $ for all $ \bm{z}_1, \, \bm{z}_2 \in \mathbb{R}^n$. 
Checking orthant monotonicity is relatively easy if $ f$ is differentiable everywhere, as it corresponds to structural balance of the graph of the Jacobian matrix of $ f$ at any point $ \bm{z} \in \mathbb{R}^n$. In particular, as stated in Proposition~\ref{prop:monotone-funct1} in the SI, a differentiable function $  f\, :  \, \mathbb{R}^n \to \mathbb{R}^n  $ is monotone w.r.t. $ \mathbb{S}$ if and only if $ S \pde{f}{\bm{z}}(\bm{z}) S \geq 0 $ $ \forall \, \bm{z} \in \mathbb{R}^n$, i.e., if and only if $ \mathcal{G}\left( \pde{f}{\bm{z}}(\bm{z}) \right)  $  is structurally balanced for some gauge matrix $ S$, $ \forall \, \bm{z} \in \mathbb{R}^n$.
To understand this result, it is enough to notice that  monotonicity with respect to the $ \mathbb{S}$ orthant corresponds to nonnegativity of the directional derivative of $ f$ along the increasing $ \mathbb{S} $ direction. 
The result then follows by a simple application of the chain rule. See SI for a formal proof.

\subsection*{Distance to balance/monotonicity: frustration}
\label{sec:dist-bal}

When a graph is not structurally balanced, then it is of interest to understand how distant it is from being structurally balanced.
One of the measures adopted in the literature for this scope is the so-called {\em frustration} \cite{facchetti2011computing,aref2018measuring}. This concept derives from Ising spin glass theory \cite{binder1986spin,mezard1987spin,mezard2009information}, and in fact it can be expressed in terms of the ``ground state energy'', i.e., the minimum over the vector of ``spins'' $ \bm{s}$ of a (normalized) energy-like functional
\beq
e(\bm{s}) =\frac{ \sum_{i,j} \left[ |A| - S AS\right]_{ij}}{ 2 \sum_{i,j} \left[ |A| \right]_{ij} },
\label{eq:energy1}
\eeq
where $A$ is the weighted adjacency matrix of the graph, $ |A| $ is its absolute value, and $\bm{s}$ is the signature vector of the aforementioned gauge diagonal matrix $ S = \diag(\bm{s})$. 
By construction it is $ 0 \leqslant e(\bm{s}) \leqslant 1 $. If $ A$ is given, the frustration index is the minimum of this cost functional over all $\bm{s}$:
\beq
\epsilon = \min_{\substack{\bm{s} =[ s_1\, \dots \,s_n] \\s_i=\pm 1}} e(\bm{s}) .
\label{eq:frustr1}
\eeq
For a connected $ \mathcal{G}(A) $, $ \epsilon =0$ iff $ \mathcal{G}(A) $ is structurally balanced. 
In fact, $ \mathcal{G}(A)$ structurally balanced corresponds to the existence of a $ S$ such that $ SAS \geq 0$. 
When instead $ \mathcal{G}(A) $ is not structurally balanced, then $ \epsilon >0$ \cite{facchetti2011computing}.

\subsection*{Heuristic algorithm for computing frustration}
\label{sec:algorithm}
For structurally unbalanced graphs, computing $ \epsilon $ is an NP-hard problem, equivalent to solving a MAX-CUT problem \cite{barahona1982computational}, or to a MAX-2XORSAT problem \cite{mezard2009information}. 
Consider the weighted adjacency matrix $A$ and its symmetrized version $ A_u = A+A^\top $. 
Let us rewrite \eqref{eq:energy1} as
\beq
e(\bm{s}) =\frac{1}{2} \left(1 - \alpha  \sum_{i,j} \left[ SA_u S\right]_{ij} \right) = \frac{1}{2}\left( 1 - \alpha \mathds{1}^\top SA_u S \mathds{1} \right),
\label{eq:compact-energy}
\eeq
where $ \mathds{1} $ is the vector of all 1s and $ \alpha = 1/ \sum_{i,j} \left[ |A_u |\right]_{ij}$. 
To search for the minimum of \eqref{eq:compact-energy}, we use the greedy heuristic described e.g. in \cite{iacono2010determining}, adapting it from adjacency matrices with binary entries $ \pm 1 $ to our weighted $A_u$.
The algorithm amounts essentially to flipping sign to the spins corresponding to rows of negative sum in $ SA_uS$, until no negative row sum remains.
See Algorithm~\ref{alg:frustration} in Section~\ref{app:algorithm} of the SI for a pseudocode.

\section*{Results}

\subsection*{Monotonicity and structural balance for DNNs}
\label{sec:main}
In this section we show how to apply the concepts of monotonicity and structural balance to DNNs.

Given an input $ \bm{x}\in \mathbb{R}^{n_0} $ and an output $ \bm{y} \in \mathbb{R}^{n_h}$, let $ \phi\,: \, \mathbb{R}^{n_0} \to \mathbb{R}^{n_h} $ be a function from input to output. 
To model $ \phi$, consider a deep neural network composed of $ h$ layers (meaning $ h-1$ hidden layers, or $ h+1 $ total layers, including input and output). Each layer $ \ell $ has $ n_\ell $ nodes ($ n_0 = $ n. of inputs, and $ n_h = $ n. of outputs). 
Let $ W^{(\ell)}$ and $ \bm{b}^{(\ell)}$ be the weights associated to the $ \ell$-th layer (of size $ n_\ell \times n_{\ell -1} $ and $ n_\ell \times 1 $ respectively). 
Denoting $ \bm{z}_\ell $ the state of the nodes of the $ \ell$-th layer, with $ \bm{z}_0 = \bm{x} $ the input and $ \bm{z}_h = \bm{y} $ the output, then the input-output function $ \phi$ describing the network can be written as 
\beq
\begin{split}
\bm{z}_{\ell } &= \sigma \left( W^{(\ell)} \bm{z}_{\ell-1} + \bm{b}^{(\ell)} \right) , \quad \ell = 1, \ldots, h-1 \\
\bm{y} & = W^{(h)} \bm{z}_{h-1} + \bm{b}^{(h)} ,
\end{split}
\label{eq:DNN1}
\eeq
where $ \sigma (\cdot ): \mathbb{R}^{n_\ell } \to  \mathbb{R}^{n_\ell } $ is a componentwise activation function.
We shall call $ \bm{q}_\ell = W^{(\ell)} \bm{z}_{\ell-1} + \bm{b}^{(\ell)} $ the preactivation state at the $\ell$-th layer.

While in \eqref{eq:DNN1} each layer is defined as contributing a matrix of weights $ W^{(\ell)}$, a vector of biases $ \bm{b}^{(\ell)}$ and an activation function $ \sigma(\cdot)$, in the practice of the machine learning literature each operation that appears in a DNN is typically denoted as a ``layer''. 
In particular, in the DNNs we consider in this study, elements such as dense connections, convolutions, residual connections, but also (max or average) pooling, ReLU, batch normalization, depth concatenation, additions, softmax, etc., are generically referred to as layers. 
Each of these layers can be mapped into the representation \eqref{eq:DNN1}, even though this sometimes requires us to overload a bit the notation of \eqref{eq:DNN1}. 
See the details in Section~\ref{app:weights} in the SI.

\paragraph{Adjacency matrix of a DNN.}

\begin{figure}
    \centering
    \includegraphics[trim=0cm 10.5cm 4.2cm 0cm,clip=true, width=\linewidth]{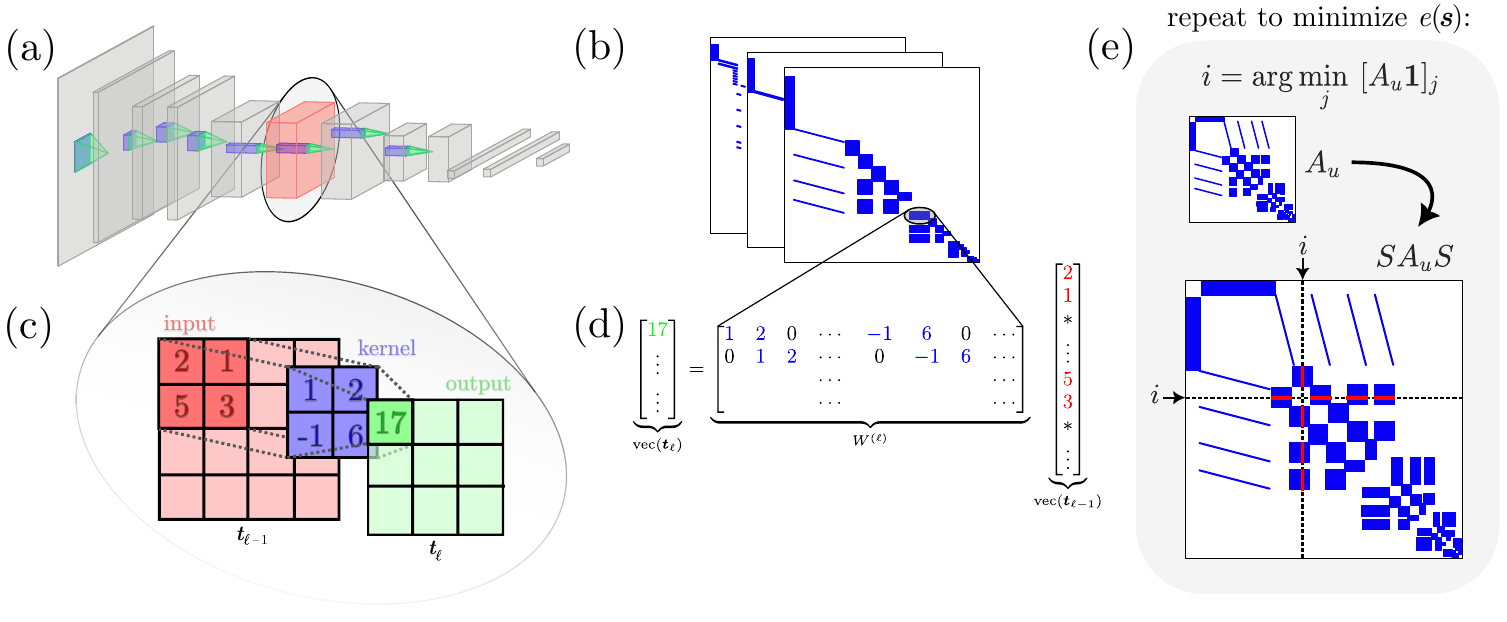}
    \caption{Constructing the adjacency matrix of a CNN. Each layer of the CNN (a) is represented by as sub-diagonal block in the adjacency matrix of the network (b). Convolutional operations (c) are represented as linear transformations (d), which give the weights of the edges in the multibanded Toeplitz adjacency matrix. (e): Heuristic minimization procedure used to compute frustration: choose the row of the adjacency matrix with most negative sum and apply a gauge transformation to it (flipping the sign of all entries in the row and column), until no negative row sum remains.}
    \label{fig:construct-dag}
\end{figure}

Topologically, a DNN like \eqref{eq:DNN1} is a DAG, whose adjacency matrix can be constructed from the weights $ W^{(\ell)}$, $ \bm{b}^{(\ell)}$.
For that, it is convenient to use a representation in homogeneous coordinates.
Denote $ \bm{z} =\begin{bmatrix} \bm{x}^\top & \bm{z}_1^\top & \ldots & \bm{z}_{h-1}^\top & \bm{y}^\top  \end{bmatrix}^\top \in \mathbb{R}^n $ the stack vector of all node states (i.e., the state of all neurons, including inputs and outputs), of dimension $ n = \sum_{\ell=0}^h n_\ell  $, and $ \bar{\bm{z}} =  \begin{bmatrix} \bm{z}^\top & 1 \end{bmatrix}^\top $ the vector of dimension $ n+1$ containing also a single constant state corresponding to the bias terms in \eqref{eq:DNN1}. For a DNN without multiple branches, where the output of layer $\ell$ acts as input only to layer $\ell+1$, its adjacency matrix is composed by a block-shift part (subdiagonal blocks in \eqref{eq:A}), plus a final column containing the bias terms:
\beq
\bar A = \left[ \begin{array}{ccccc|c} 
		0 & 0 & 0 &  \hdots & 0  & 0 \\ 
		W^{(1)} & 0 & 0 & \hdots & 0 & \bm{b}^{(1)} \\ 
		0 & W^{(2)} & 0 &  \hdots & 0  & \bm{b}^{(2)} \\ 
		\vdots & \vdots  & \ddots & \ddots & \vdots & \vdots \\ 
		0 & 0 &  \hdots & W^{(h)} & 0 & \bm{b}^{(h)} \\
        \hline 
		0 & 0 & \hdots & 0 & 0 & 1
	\end{array} \right]  = \left[ \begin{array}{c|c} A & \bm{b} \\ \hline 0 & 1 \end{array}\right] \in \mathbb{R}^{(n+1) \times (n+1)} 
 \label{eq:A}
\eeq
where $ A\in \mathbb{R}^{n \times n} $ is the submatrix of $ \bar A $ obtained dropping the last row and column and $ \bm{b} =\begin{bmatrix} 0 & (\bm{b}^{(1)})^\top & \ldots (\bm{b}^{(h)})^\top   \end{bmatrix}^\top \in \mathbb{R}^n $
(see sketch in Fig.~\ref{fig:construct-dag} for the case of a CNN).
When the network has multiple branches, as is common when using residual connections, or when concatenating or adding the output of layers, then also blocks in the lower triangular part of $A$ in \eqref{eq:A} can be non-zero. 
For compactness of notation, we omit here these types of connections. All results can be extended to include them at the cost of some extra bookkeeping, see Section~\ref{app:weights} in the SI.

\paragraph{A compact description of the DNN based on the hidden states.}

On a similar note, to streamline the analysis we can also express the various nonlinearities of the DNN (such as ReLU, max-pooling, batch normalization) as special cases of the activation function $ \sigma(\cdot)$  (with a slight abuse of notation, see details in Section~\ref{app:weights} of the SI). 
Since $ \sigma$ acts componentwise, using $ A $ and $\bm{b}$ the DNN \eqref{eq:DNN1} can be vectorized as follows
\beq
\bm{z} = \begin{bmatrix} \bm{z}_0 \\ \bm{z}_1 \\ \vdots \\ \bm{z}_{h-1} \\ \bm{z}_h \end{bmatrix}
= \begin{bmatrix} \bm{x} \\ \sigma \left( W^{(1)} \bm{z}_0 + \bm{b}^{(1)} \right) \\ \vdots \\ \sigma \left( W^{(h-1)} \bm{z}_{h-2} + \bm{b}^{(h-1)} \right) \\   W^{(h)} \bm{z}_{h-1} + \bm{b}^{(h)} \end{bmatrix} = f \left( A \bm{z} + \bm{b} \right)
\label{eq:DNN2}
\eeq
where $ f : \mathbb{R}^n \to \mathbb{R}^n $ indicates a componentwise map gathering all (vectorized) activation functions $ \sigma$ (plus the trivial map on the first $ n_0 $ components). 
If the parameters $A$ and $ \bm{b}$ are given, then $ f $ is a map from $ \bm{z} $ to itself. 
For later reference, denote $ \bm{\theta} $ the stack vector of all (vectorized) $ W^{(\ell)}$, and $ \bar{\bm{\theta}} $ the vector containing $ \bm{\theta} $ and $ \bm{b}$.

\paragraph{Structural balance and monotonicity for DNNs.}
\label{sec:struc-bal-mon-DNN}

An easy way to investigate monotonicity of the DNN \eqref{eq:DNN2} is to consider its Jacobian matrix.
Observe first that activation functions $ \sigma(\bm{q})$ such as ReLU, pooling and normalization are all characterized by $ \pde{\sigma(\bm{q})}{\bm{q}} \geq0$, even when the derivative should be replaced by a generalized derivative and the inequality by an inclusion because of the non-differentiability of $ \sigma$.

When we restrict to these activation functions, checking monotonicity of the DNN map $ f$ is rather straightforward, as it amounts to checking structural balance of the $ n \times n$ subgraph $ \mathcal{G}(A)$ of  $ \mathcal{G}(\bar A)$.
As shown in Theorem~\ref{thm:monot-bal} in the SI, the DNN map $ f $ is monotone if and only if $ \mathcal{G}(A) $ is structurally balanced. 
In fact, nonnegativity and ``separability'' (i.e., diagonal structure) of the partial derivatives of $ \sigma$ imply that the sign pattern of the Jacobian of $ f$ reflects that of $A$. 
More precisely, if we denote $ F(\bm{z})= \pde{f}{\bm{z}}({A\bm{z}+\bm{b}}) $ the Jacobian of $ f$ at $ \bm{z}$, it is 
\beq
\sign(F(\bm{z}) ) = \sign(\diag(\mathbb{I}(\bm{z})) A),
\label{eq:sign-Jac}
\eeq
where $ \sign(\cdot ) $ is the sign function, $ \mathbb{I}(\bm{z} ) $ is a vectorized indicator function (here equal to 1 at a node if the associated $ z_i $ is active, equal to 0 otherwise), and $ \diag(\cdot) $ is the diagonal matrix, see Lemma~\ref{lemm:Jac} in the SI.
In words, the adjacency matrix $A$ determines the signature of the Jacobian of the DNN, modulo some vanishing rows in correspondence of hidden nodes that are inactive.
The value of $ \mathbb{I}(\bm{z}) $ depends on the input $ \bm{x}$ and on the type of activation function $ \sigma (\cdot) $ present at a node, see Section~\ref{app:active-subn} in the SI for more details.

\paragraph{Active subnetwork for an input.}
Notice that when a node is inactive then also its outgoing edges become irrelevant for the DNN map. 
Hence, for a specific input $ \bm{x}$, in place of \eqref{eq:sign-Jac} one can consider the more compact active sub-adjacency matrix
\[
A_{\rm act} (\bm{x})   =  {\rm diag}( \mathbb{I} ( \bm{z}))\, A \, {\rm diag}( \mathbb{I} ( \bm{z}) ).
\]
Similarly to \eqref{eq:frustr1}, we can compute the frustration index of $ A_{\rm act} $ as 
\beq
\epsilon_{\rm act}(\bm{x})  = \min_{\substack{\bm{s} =[ s_1\, \dots \,s_n] \\s_i=\pm 1}}
\frac{ \sum_{i,j} \left[ |A_{\rm act}(\bm{x}) | - S A_{\rm act}(\bm{x}) S\right]_{ij}}{ 2 \sum_{i,j} \left[ |A_{\rm act}(\bm{x}) | \right]_{ij} }.
\label{eq:frustr_act}
\eeq

\paragraph{Relation with IO monotonicity.}
\label{sec:IO-monot}

The concept of monotonicity considered in this paper differs from the one normally used in the DNN literature \cite{cano2019monotonic,liu2020certified,runje2023constrained,sharma2020testing,sill1997monotonic}, which in our setting could be called Input-Output (IO) monotonicity. 
A DNN, represented as an IO map $ \bm{y} = \phi(\bm{x}, \bar{\bm{\theta}} )$, is IO monotone with respect to the input order $ \mathbb{S}_{\bm{x}} $ and the output order $ \mathbb{S}_{\bm{y}}$ if $  \forall \, \bm{x}_1, \, \bm{x}_2 $
\beq
  \bm{x}_1 \leq_{\mathbb{S}_{\bm{x}}} \bm{x}_2\; \Rightarrow  \; \phi(\bm{x}_1, \bar{\bm{\theta}} ) \leq_{\mathbb{S}_{\bm{y}}} \phi(\bm{x}_2, \bar{\bm{\theta}} ).
 \label{eq:IOmonot1}
 \eeq
Notice that since in this case the input and output spaces are different, two distinct partial orders  $ \mathbb{S}_{\bm{x}} $ and  $ \mathbb{S}_{\bm{y}}$ must be specified in the definition of IO monotone.
Since $ \bm{x}$ and $ \bm{y}$ are parts of the ``complete state'' $ \bm{z} $, both  signature vectors $ \bm{s}_{\bm{x}} $ and $ \bm{s}_{\bm{y}} $ associated to $\mathbb{S}_{\bm{x}} $ and  $ \mathbb{S}_{\bm{y}}$ are subvectors of the signature $ \bm{s} $ associated to $ \mathbb{S} $. 

A key result that we rely upon in this paper is that structural balance of the DAG is a sufficient condition for IO monotonicity. In fact, from \eqref{eq:sign-Jac}, the sign pattern of the Jacobian of $ \phi$ is inherited from that of $A$, hence if the graph $ \mathcal{G}(A)$ is structurally balanced then the IO map $ \bm{y} = \phi(\bm{x}, \bar{\bm{\theta}} )$ must be IO monotone, see Theorem~\ref{thm:IOmonot-bal} in the SI for a detailed proof. 
Consequently, if the DNN map $ f $ is monotone then the IO map $ \bm{y} = \phi(\bm{x}, \bar{\bm{\theta}} )$ is IO monotone (Corollary~\ref{cor:IOmonot-monot} in the SI).

The crucial advantage of dealing with ``hidden state monotonicity'', i.e., with the map $ f$ instead of the IO map $ \phi$, is that we obtain a notion which gives insight into the organization of the DNN induced by the training algorithm, which the simpler IO perspective cannot account for.

\subsection*{Quantifying near-monotonicity on DNNs}
If a DNN is not monotone, then \eqref{eq:IOmonot1} cannot be obeyed rigorously for all $ \bm{x}_1, \, \bm{x}_2 $, and we also know from Theorem~\ref{thm:monot-bal} in the SI that the graph $ \mathcal{G}(A)$ associated to the DNN cannot be structurally balanced, i.e., $ \epsilon >0$. 
As mentioned in the Methods, the frustration $ \epsilon$, a measure of distance to structural balance, is often taken as a proxy for the distance to monotonicity of a system whose Jacobian has graph $ \mathcal{G}(A)$. 
However, $ \epsilon$ provides only an indirect measure.
In this section we aim to compute a more direct measure of near-monotonicity.
The rationale is that when the frustration $\epsilon$ is low, one would expect that to some extent the partial order determined by $\mathbb{S}_{\bm{x}} $ and $ \mathbb{S}_{\bm{y}} $ in the IO map $ \phi(\cdot) $ is still respected. One possible way to quantify this is to count how many output nodes $ y_i = \phi_i (\bm{x}, \bar{\theta}) $, $ i=1, \ldots, n_h$, still respect the output order $  \mathbb{S}_{\bm{y}} $ when $  \bm{x}_1 $ and $\bm{x}_2 $   respect the input order, i.e., for $ \bm{x}_1 \leq_{\mathbb{S}_{\bm{x}}} \bm{x}_2 $. 
Before embarking in this quantification, in next section we explain why for a non-monotone function, even a near-monotone one, it is a priori impossible to disambiguate increasing vs decreasing near-monotonicity w.r.t. the partial order, just by looking at the partial order pair $ \{ \mathbb{S}_{\bm{x}}, \, \mathbb{S}_{\bm{y}} \}$ and at $ \epsilon$. 

\paragraph{Decreasing/increasing trends in near-monotonicity: an heuristic explanation of an intrinsic ambiguity.}

Lack of exact monotonicity introduces an ambiguity in the near-monotonicity pattern, which can be understood by looking at IO paths, i.e., at directed paths on $ \mathcal{G}(A)$ connecting an input $ i$ to an output $ j$. 
Once we drop the edge direction, any two equal-ends IO paths, say $ p_{ij}^1 $ and $ p_{ij}^2 $, form an undirected cycle, which is necessarily positive when the DNN is monotone, even though it could be $ p_{ij}^k>0$ or $ p_{ij}^k<0 $, $ k=1,2$ (all equal-ends IO paths between $ i$ and $ j$ have to have the same sign in the monotone case, otherwise negative undirected cycles would appear). 
However, when the DNN is not monotone, then some negative undirected cycles exist, implying that equal-ends IO paths with opposite signs are present. 
IO paths of opposite signs have opposite effects on the output order: if a positive IO path reflects an expression like \eqref{eq:IOmonot1} (``increasing output'' in the partial order $ \mathbb{S}_{\bm{y}}$), a negative IO path reflects instead its opposite $ \phi(\bm{x}_1, \bar{\bm{\theta}} ) \geq_{\mathbb{S}_{\bm{y}}} \phi(\bm{x}_2, \bar{\bm{\theta}} )$ (``decreasing output'').
For given $ \bm{x}_1 $ and $  \bm{x}_2 $ obeying $ \bm{x}_1 \leq_{\mathbb{S}_{\bm{x}}} \bm{x}_2 $, the two cases coexist when a network is not monotone. 
Frustration alone cannot predict the prevalence of positive or negative IO paths, meaning that predicting whether the near-monotonicity is of increasing or decreasing type cannot be done a priori, and for a given DNN it can vary with $ \bm{x}_1 $ and $  \bm{x}_2 $.

\paragraph{Quantifying near-monotonicity in DNNs: a direct test.}

Let $ \Omega $ be the output alignment fraction for increasing monotonicity: 
$ \Omega :=   \frac{1}{n_h} \mathbb{I} \left( \phi(\bm{x}_1, \bar{\bm{\theta}}) \leq_{\mathbb{S}_{\bm{y}}} \phi(\bm{x}_2,\bar{\bm{\theta}}) \right)$, where $ n_h $ is the output size, i.e., $ \Omega $ is the fraction of outputs for which $ \phi(\bm{x}_1, \bar{\bm{\theta}}) \leq_{\mathbb{S}_{\bm{y}}} \phi(\bm{x}_2,\bar{\bm{\theta}}) $ holds.
Then we say that $ \phi$ is {\em $ \lambda$-monotone} with respect to the IO partial order pair $ \{  \mathbb{S}_{\bm{x}} ,  \mathbb{S}_{\bm{y}} \}$ if 
\beq
  \bm{x}_1 \leq_{\mathbb{S}_{\bm{x}}} \bm{x}_2\; \Rightarrow  \;  
P \left( \lvert \Omega - 0.5 \rvert \geq \lambda \right) \geq 2\lambda.
 \label{eq:Iomonot2}
\eeq
The two (mutually exclusive) inequalities, $ \Omega > 0.5 +\lambda $ and $ \Omega < 0.5 -\lambda$, specify the size of $ \Omega $ in the two possible cases resulting from the increasing/decreasing ambiguity just described.
Eq.~\eqref{eq:Iomonot2} expresses the probability at which these two inequalities are expected to occur when an image $\bm{x}_1$ is perturbed in an increasing direction with respect to $\mathbb{S}_{\bm{x}}$.
The value $ \lambda\in [0, \, 0.5]$ can be taken as a measure of (increasing or decreasing) IO monotonicity associated to an input $ \bm{x}_1 $:
the larger $ \lambda$ is, the closer to IO monotone the DNN is.
$ \lambda=0 $ means no monotonicity at all, while 
$ \lambda=0.5$ means exact IO monotonicity, and corresponds to the ``lower'' (decreasing) condition in \eqref{eq:Iomonot2} becoming void (no increasing/decreasing ambiguity is present in the exact monotone case, which is always increasing as in \eqref{eq:IOmonot1}).

\subsection*{Numerical experiments}
\label{sec:numerical}
To quantify the frustration and the degree of monotonicity in state-of-the-art DNNs,
we consider seven pre-trained networks available in \textsc{Matlab}'s Deep Learning Toolbox (ShuffleNet \cite{zhang2018shufflenet}, SqueezeNet \cite{iandola2016squeezenet}, ResNet18 \cite{he2016deep}, AlexNet \cite{krizhevsky2017imagenet}, and GoogLeNet \cite{szegedy2015going}) and from PyTorch Hub (AlexNet and GoogLeNet), which are all convolutional-type neural networks trained on the ImageNet dataset \cite{deng2009imagenet}.

\paragraph{Frustration.}

We construct the adjacency matrix of the CNNs as described above, restricting to the homogeneous matrix $A$ and disregarding the bias terms $ \bm{b}$, see details in Section~\ref{app:weights} and Fig.~\ref{fig:pretrained_adjmat} of the SI. We then run Algorithm \ref{alg:frustration} (see Section~\ref{app:algorithm} of the SI) to estimate the frustration index \eqref{eq:frustr1} in correspondence of the real weights of the networks.
To explore the energy landscape associated to these weights, we repeat the calculation of frustration 80 times, initializing the sign pattern matrix $ S $ differently each time, in particular applying 1 million random single-neuron gauge transformations at the start of the algorithm (we explore in this way 80 different replicas of the same ``quenched'' spin glass).
The least among these 80 values of frustration is our best estimate of the frustration of a CNN.
To assess if this frustration value is below or above our expectations, we compare it with the frustration obtained from several null models, in which the network weights have been reshuffled or randomized. The first null model (N1) is constructed by randomly reordering the position of the parameters in each convolutional layer and in each dense layer. The reshuffling occurs within each layer and, for the convolutional layers, all repeated instances of a parameter in the Toeplitz matrix $ W^{(\ell)}$ are replaced with another parameter but they remain identical. 
Accordingly, the recomputed adjacency matrix has the same topology and convolutional structure as the original one. 
The second null model (N2) is obtained by shuffling the weights of the adjacency network across all convolutional and dense layers, so that the new adjacency matrix has entries in the same positions as the real network and N1 model but it no longer represents convolutions. A third null model (N3) is constructed by drawing new weights for the convolutional layers and dense layers from either uniform (Xavier) or normal (He) distributions to simulate the state of a CNN before training starts.
The pooling layer weights are not modified in any null model.
We create 80 null models of each kind and explore their energy landscape by applying our algorithm to two instances of each null model, corresponding to different initial $S$ (the second instance is obtained randomizing the initial $S$ through 1 million random single-neuron gauge transformations). 

\begin{figure}
    \centering
    \includegraphics[trim=0.5cm 9.6cm 0.5cm 9.5cm, clip=true, width=\linewidth]{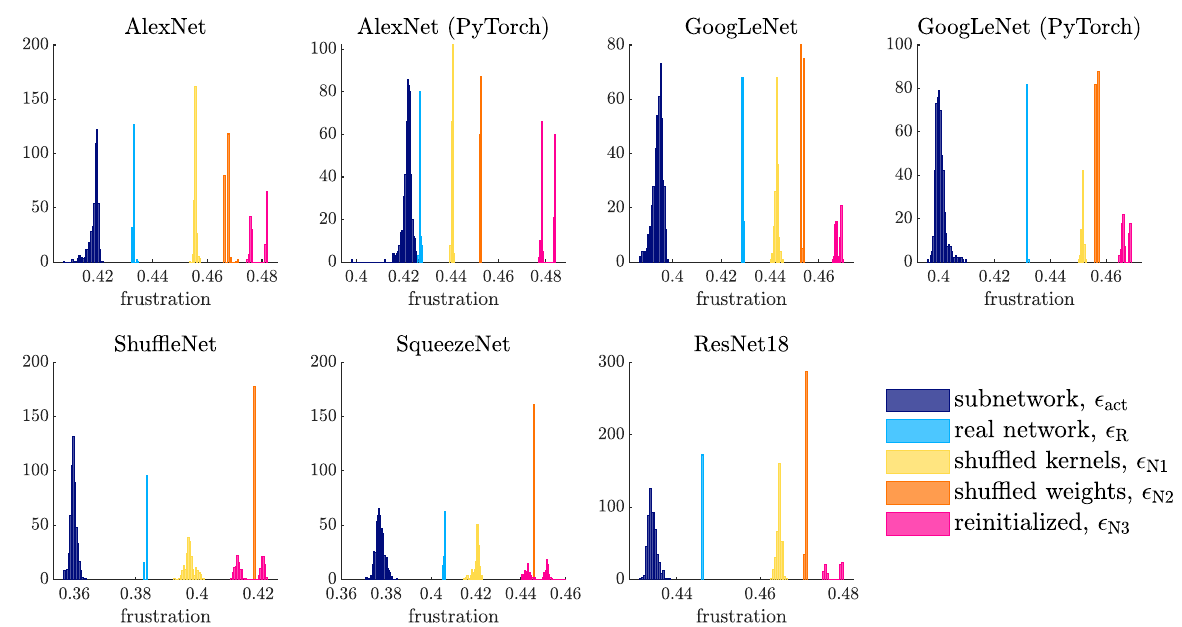}
    \caption{ Frustration of the seven pretrained CNNs ($ \epsilon_{\rm R}$), their active subnetworks ($\epsilon_{\rm act}$) and their null models ($ \epsilon_{\rm N1}$, $ \epsilon_{\rm N2}$ and $ \epsilon_{\rm N3}$).}
    \label{fig:simulations}
\end{figure}

The results of these computations are shown in Fig.~\ref{fig:simulations}.
Denote $ \epsilon_{\rm R}$ the frustration for the network with the real parameters, and $  \epsilon_{\rm N1}$, $  \epsilon_{\rm N2} $ and $  \epsilon_{\rm N3}$ the frustration of the three null models. 
We always find $ \epsilon_{\rm R}<   \epsilon_{\rm N1} <  \epsilon_{\rm N2} $ and the difference, though small, is statistically significant: Welch's t-test shows significant difference in distribution means between $ \epsilon_{\rm R}$, $   \epsilon_{\rm N1} $ and $  \epsilon_{\rm N2} $ at arbitrarily low p-value ($<10^{-50}$). 
Recall that $ 0 \leq \epsilon \leq 1 $. A completely random network, with random weights, would corresponds to $  \epsilon \approx 0.5$. 
This is roughly what happens to $ \epsilon_{\rm N3}$ in the reinitialized model N3, where 
the two distinct peaks appearing in some of the networks correspond to the two different initializations mentioned above, with He initialization tending to give slightly lower frustration. 
For most CNNs, some decrease in frustration is already visible in $ \epsilon_{\rm N2}$ where the reshuffling of the real weights destroys the convolutional structure, but in $A$ many identical weights are repeated, which already favors a reduction in disorder. 
When the convolution structure is preserved, as in the $ {\rm N1} $ null model, this effect is enhanced, and in fact $  \epsilon_{\rm N1}$ is significantly smaller than $ \epsilon_{\rm N2}$ and $ \epsilon_{\rm N3}$. 
The optimized real weights reduce even further the frustration: in all cases $ \epsilon_{\rm R}$ is significantly smaller than the frustration of the null models.

In addition, Fig.~\ref{fig:simulations} shows that all 80 values of $ \epsilon_{\rm R}$ are always rather close to each other: the histograms of $ \epsilon_{\rm R}$ have always a limited dispersion. This means that the local minima reached by the algorithm are all close to each other in value, i.e., that the energy landscape of the spin glass is not particularly rugged around its ground states.

As the results are similar for all the CNNs we consider, even for those implemented and trained in a different platform, it appears that they hold regardless of specific details such as the values of the hyperparameters chosen for the training and other implementation technicalities.

Also for the active subnetwork $ A_{\rm act}(\bm{x})$ associated to each image $ \bm{x}$, it is possible to compute the frustration $ \epsilon_{\rm act}$ using \eqref{eq:frustr_act}. 
For this purpose, we use images from the Imagenette subset of the ImageNet dataset (see Section~\ref{app:data} in the SI), which contains 10 classes of images. For around 500 images from these 10 classes the frustration $ \epsilon_{\rm act}$ is shown in Fig.~\ref{fig:simulations}. In all networks it is nearly always $ \epsilon_{\rm act} < \epsilon_{\rm R}$, i.e., the adjacency matrix $A$ is organized in such a manner that each active subnetwork $ A_{\rm act}(\bm{x})$ is even less frustrated than the entire $A$.

\begin{figure}
    \centering
    \includegraphics[trim=0cm 9.4cm 0cm 9.3cm, clip=true, width=\linewidth]{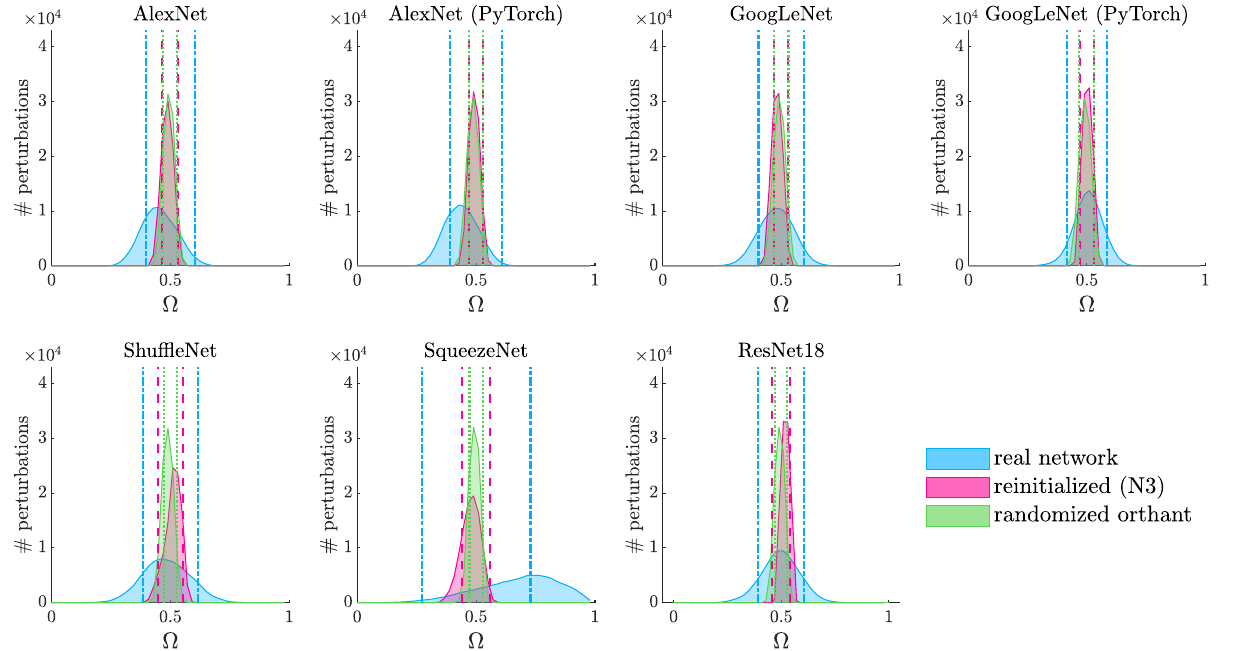}
    \caption{Histograms of the output alignment fraction $ \Omega$ and associated values of $ \lambda $, for the real networks (azure) and for two null models (violet and green).
    For each of the seven CNNs, the histogram of the real network differ significantly from the other two: it is much broader on both sides of $ 0.5$, showing that $ |\Omega -0.5| $ is much larger than expected from null models. For each histogram, the two vertical bars identify the interval of width $ 2 \lambda $ described in Eq.~\eqref{eq:Iomonot2}. See Fig.~\ref{fig:lambda-panel} in the SI for details on how to compute $ \lambda$.
}
    \label{fig:Omega-panel-alt}
\end{figure}

\begin{figure}
    \centering
    \includegraphics[trim=0cm 9.4cm 0cm 9.3cm, clip=true, width=\linewidth]{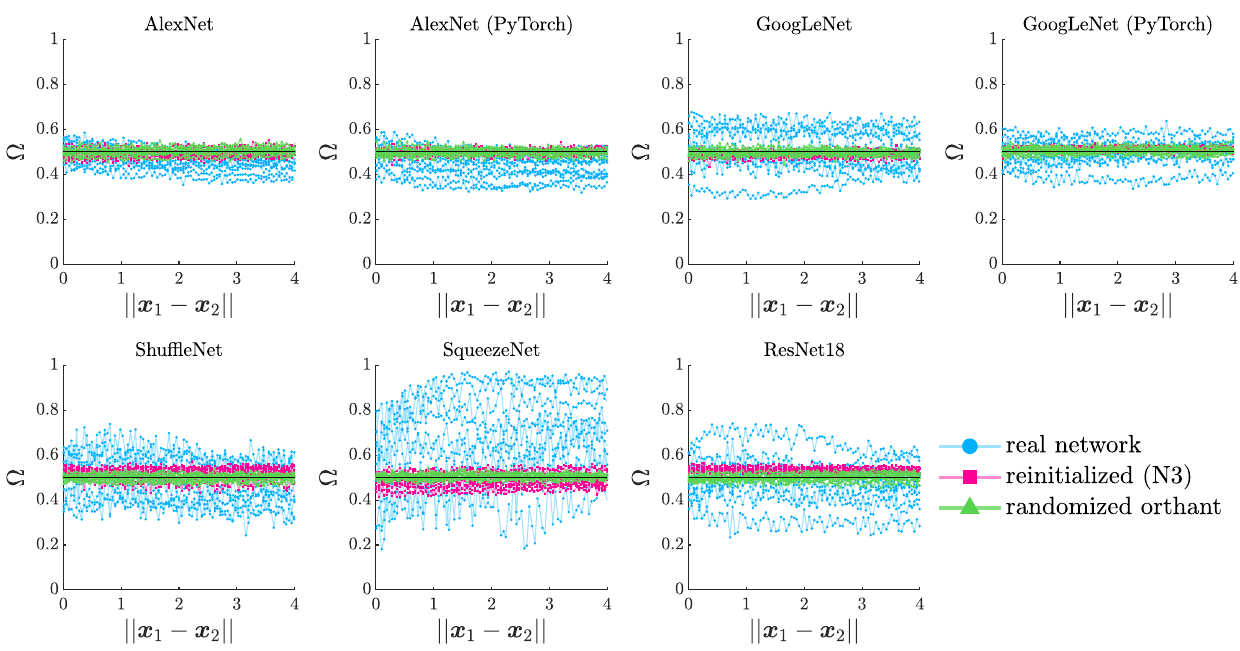}
    \caption{Output alignment fraction $ \Omega$ in response to 100 perturbations $\bm{\delta}_{\mathbb{S}_{\bm{x}}}$ of different amplitude for a few images $\bm{x}_1$, as a function of the amplitude $ \|\bm{\delta}_{\mathbb{S}_{\bm{x}}} \|$, for each of the seven CNNs. For the real network (azure) the alignment is normally larger than for the two null models (violet and green), and all perturbations  for an image (linked by a continuous line) tend to be on the same side of $ 0.5$, i.e., always increasing or always decreasing.
    }
    \label{fig:Omega-scatter-panel}
\end{figure}

\paragraph{Near-monotonicity.}

\begin{figure}
    \centering
    \includegraphics[trim=0cm 10.5cm 0cm 10.5cm, clip=true, width=0.85\linewidth]{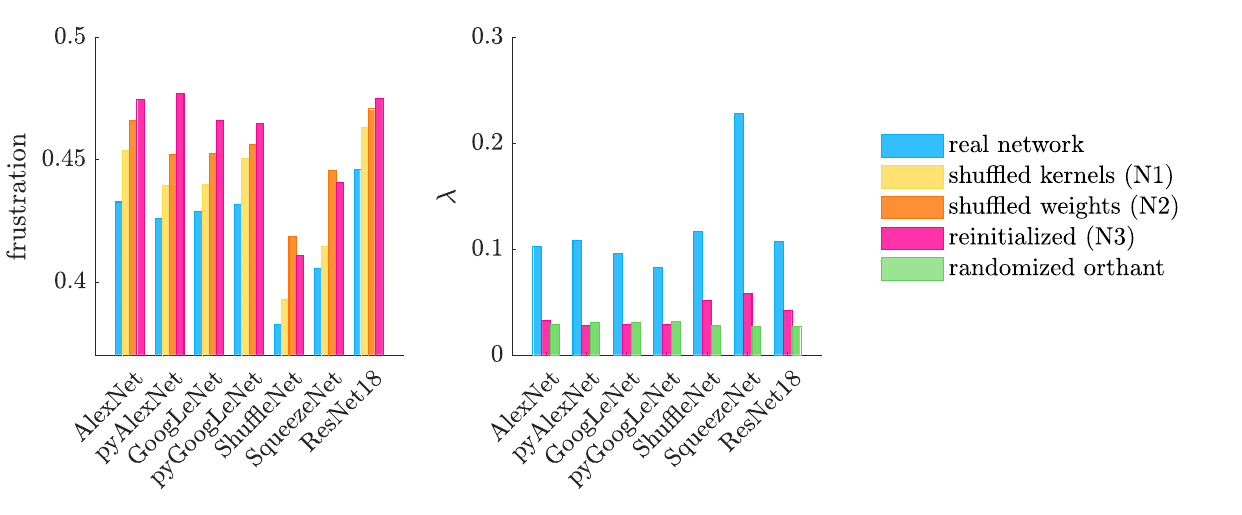}
    \caption{Summary of the results. (a) Frustration computed for all real networks and null models. (b) $\lambda$-monotonicity values for all networks with real weights, re-initialized weights (N3) and real weights but with input perturbation in a random direction. }
    \label{fig:lambda-frustr-compare}
\end{figure}

To check whether the reduced frustration results in a CNN which is near-monotone, i.e. in an IO function $ \bm{y} =\phi(\bm{x}, \bar{\bm{\theta}})$ which is closer to monotone than expected, given an image $ \bm{x}_1$, we perturb it along the direction indicated by the input gauge transformation $ S_{\bm{x}}$, i.e., we add a randomly chosen perturbation $ \bm{\delta}_{\mathbb{S}_{\bm{x}}} = S_{\bm{x}} \bm{\delta} \in \mathbb{S}_{\bm{x}}$, with $ \bm{\delta} \in \mathbb{R}^{n_0} $, $\bm{\delta} > 0 $,  so that $ \bm{x}_2 = \bm{x}_1 + \bm{\delta}_{\mathbb{S}_{\bm{x}}} $ is s.t. $ \bm{x}_2 \geq_{\mathbb{S}_{\bm{x}}} \bm{x}_1 $. 
Each element of $ \bm{\delta} $ is drawn from a uniform distribution.
We then compute the associated logit (pre-softmax) 
outputs $ \bm{y}_k = \phi(\bm{x}_k, \bar{\bm{\theta}})$, $ k=1, 2 $, and calculate the output alignment fraction $ \Omega$. 
For each image we repeat this calculation 100 times, changing the magnitude $ \| \bm{\delta}\|$ of the random perturbation. 
The procedure is repeated for 1000 images on each of the seven  CNNs.
The resulting histograms for $ \Omega$ are shown in Fig.~\ref{fig:Omega-panel-alt}. We use them to compute $ \lambda $ that satisfies \eqref{eq:Iomonot2}, see  Fig.~\ref{fig:lambda-panel} in the SI. 
For all CNNs, $ \lambda $ is in the range $ 0.1\div 0.25$, and always significantly larger than when computed on null models. 
In this case we consider as null models N3\footnote{N1 and N2 sometimes lead to vanishing -- or exploding -- hidden states, hence they are not suitable to compute outputs.} and a randomly chosen partial order in place of the $ \{ \mathbb{S}_{\bm{x}}, \, \mathbb{S}_{\bm{y}}\}$ determined by Algorithm~\ref{alg:frustration}.
For all seven CNNs, the histograms of output alignment  $ \Omega$ associated to these two null models are much more concentrated around 0.5, see Fig.~\ref{fig:Omega-panel-alt}, and the corresponding $ \lambda$ almost never exceeds 0.05, see summary in Fig.~\ref{fig:lambda-frustr-compare}. 
As can be seen in Fig.~\ref{fig:Omega-panel-alt}, for the majority of CNNs (except SqueezeNet and the two AlexNet), both the cases $ \Omega >0.5 + \lambda $ (increasing monotonicity) and $ \Omega<0.5 -\lambda $ (decreasing monotonicity) are present. 
What is remarkable is that for a given image all perturbations we apply tend to induce always increasing monotonicity or always decreasing monotonicity even when the amplitude of $ \bm{\delta}$ varies, see Fig.~\ref{fig:Omega-scatter-panel} and Fig.~\ref{fig:frac-above05} in the SI. 
In addition, when for instance $ \| \bm{\delta} \| = 4 $, the perturbed image is predicted to belong to the same class as the original image in about 90\% of the cases.
Both observations confirm that the partial order direction $ \{ \mathbb{S}_{\bm{x}}, \, \mathbb{S}_{\bm{y}}\}$ determined by our computation of frustration is indeed special for the CNN, and hence that the CNNs have a near-monotone behavior along them.

\section*{Discussion}
\label{sec:interpret}
For all pretrained CNNs we consider in this study, the frustration $ \epsilon_{\rm R}$ is significantly less than expected from any kind of null model associated to the network. 
This can be interpreted by saying that a trained DNN has some degree of order and it is not a completely disordered system. 
Having some degree of order means that in a trained CNN the function $ f$ is closer to being a monotone function than a random function.

This does not mean that the CNN is monotone. In fact a value of $ \epsilon_{\rm R}\approx 0.4 $ is rather far from structural balance, and hence the CNN is rather far from behaving like a ``trivial'' (i.e., completely predictable) function as an exactly monotone function would be. Yet, a trained network has some amount of predictability which a non-trained network does not have, and the partial order we compute indeed provides a direction in which this predictability becomes manifest.
It is tempting to speculate that this reduction in frustration (or increase in monotonicity) reflects the inference power of the CNN.

An interesting observation is that also the null model N1 is less frustrated than a completely random model: $ \epsilon_{\rm N1} < \epsilon_{\rm N2}, \epsilon_{\rm N3}$. This suggests that part of the order in the CNN is due to its organization into convolutional structures, and in fact, CNNs are a {\em de facto} standard in modern DNNs. If we look at what happens in the inner organization of $A$, the Toeplitz structure associated to a convolutional layer $ W^{(\ell)}$ is such that many rows and columns of $ W^{(\ell)}$ have identical row/column sum.
It is not yet clear to us why this should favor order and be associated to lower frustration, especially in view of the fact that the reinitialized null model N3 also consists of convolutional layers and hence shares all features due to the Toeplitz structure.

What is clear is that for the set of pretrained parameters $ \bm{\theta}$ the convolutional structure leads to a less rugged landscape in the energy functional $ e(\bm{s})$.
In fact, when applying Algorithm~\ref{alg:frustration}, the rows/columns of $A$ with identical sum all flip sign simultaneously, simplifying the search for a ground state in the energy $ e(\bm{s})$. 
While this per se does not provide any information on the training dynamics, it may help explaining why even simple training algorithms are typically successful.

The computation of frustration is instrumental for the calculation of IO monotonicity, since it provides as a byproduct the partial order pair $\{ \mathbb{S}_{\bm{x}},  \mathbb{S}_{\bm{y}} \} $. 
The presence of a preferential partial order direction sheds  light on the internal organization of the CNNs: images perturbed along the input partial order direction $\mathbb{S}_{\bm{x}}$ are consistently still classified in the same class as the original image.
This suggests that the partial order direction determined by the pair $\{ \mathbb{S}_{\bm{x}},  \mathbb{S}_{\bm{y}} \} $ is robust to perturbations, hence that the CNN is unlikely to be  affected by practices like adversarial training along it.
Notice that the perturbation $ \bm{\delta}$ can have different amplitude on different pixels, provided that it $ \delta_i >0$ for all $i$ (or $ \delta_i <0$ for all $i$), so that $ \bm{\delta}_{\mathbb{S}_{\bm{x}}} $ is increasing (or decreasing) in the desired input partial order. 
As soon as the direction of the perturbation departs from $ \mathbb{S}_{\bm{x}}$, then any sign of monotonicity quickly disappears.

For DNNs, the search for implicit regularization principles, i.e., for mathematical rules underpinning the inherent bias of DNNs towards ``low-complexity'' solutions that generalize well is characterized by only limited successes, in spite of a intense research \cite{ji2019implicit,neyshabur2017exploring,ravi2024implicit,soudry2018implicit,vardi2023implicit}.

One strong form of organization that has been recently detected, the so-called neural collapse, occurs in the terminal phase of training, i.e., in the extra training phase that follows the achievement of the zero training error \cite{papyan2020prevalence}. 
Relating this form of order with our near-monotonicity is at the moment difficult, as our networks are not trained in the terminal phase, and our output (pre softmax, but post dense layer) does not coincide with the (pre dense layer) last layer studied in \cite{papyan2020prevalence}. We leave investigating the monotonicity of the terminal phase of training to the future research. 

To summarize, what we are suggesting in this paper is that near-monotonicity could be one mathematical notion characterizing implicit regularization in DNNs, especially since it is grounded on well-established ideas like the notion of disorder of Ising spin glasses. Furthermore, since frustration is representable by the valleys of an energy functional which is different from the loss function used in the training, a question that can be posed is whether frustration can be taken as a measure of complexity, to be used to understand generalizability \cite{neyshabur2017exploring,zhang2021understanding}. Also in this case, a further analysis is needed before drawing any conclusion.

%\bibliographystyle{abbrv}
%{
% \bibliography{References}}

\newpage

\appendix 

\begin{center}
{\bf \LARGE Supplementary Information}
\end{center}

\section{Mathematical results}
\label{sec:proofs}

In this section we provide mathematical formulations and formal proofs of the properties mentioned and exploited in the main text of the paper. 
In the section ``Monotonicity'' of the Methods we use the following.

\begin{proposition}
\label{prop:monotone-funct1}
A differentiable function $  f\, :  \, \mathbb{R}^n \to \mathbb{R}^n  $ is monotone w.r.t. $ \mathbb{S}$ iff $ S \pde{f}{\bm{z}}(\bm{z}) S \geq 0 $ $ \forall \, \bm{z} \in \mathbb{R}^n$, i.e., iff $ \mathcal{G}\left( \pde{f}{\bm{z}}(\bm{z}) \right)  $  is structurally balanced with gauge matrix $ S$, $ \forall \, \bm{z} \in \mathbb{R}^n$.
\end{proposition}

\paragraph{Proof of Proposition \ref{prop:monotone-funct1}.} 

If $ f $ is $ \mathbb{S}$-monotone, the partial order $ \leq_\mathbb{S} $ can be simplified by a gauge transformation. Namely, there exists $S$ such that $S \bm{z}_1 \leq S \bm{z}_2 \implies  f(\bm{z}_1) \leq_\mathbb{S}  f(\bm{z}_2) $, or, equivalently, $ S f(\bm{z}_1) \leq S f(\bm{z}_2) $.
Consider the change of basis $ \bm{r} = S \bm{z} $, which is simply a flip of the order in some orthants.
The partial order in the new basis is the usual $ \mathbb{R}^n $ order: $ \bm{r}_1 \leq \bm{r}_2 $ $ \Longrightarrow $ $ f(\bm{r}_1) \leq f(\bm{r}_2)$ $ \forall \; \bm{r}_1, \, \bm{r}_2 \in \mathbb{R}^n$. 
In the new basis, $ f$ is monotone in this usual sense. When we differentiate it along the direction $ \bm{r}_2 -\bm{r}_1 \geq 0 $, we get, using the chain rule,
\beq
\frac{d}{dt} f(\bm{r}_1 + t (\bm{r}_2 -\bm{r}_1) ) = \pde{f}{\bm{r}} (\bm{r}_1 + t (\bm{r}_2 - \bm{r}_1)) (\bm{r}_2 - \bm{r}_1) \geq 0\qquad \forall \, t\in [0, \, 1]
\label{eq:monot-df}
\eeq
By contradiction, assume that $ \left[ \pde{f}{\bm{r}} (\bm{r}) \right]_{ij} <0 $ for some index $ i, j$ and for some $ \bm{r} \in \mathbb{R}^n$. Then there exists $ \bm{r}_1 , \, \bm{r}_2 $ and $ t\in [0, \, 1]$, with $ \bm{r}_2 - \bm{r}_1 \geq 0 $, s.t. $ \bm{r} = \bm{r}_1 + t (\bm{r}_2 -\bm{r}_1) $ and \eqref{eq:monot-df} is violated. 
One example is $ \bm{r}_2 - \bm{r}_1 = \bm{e}_j $, the elementary vector having $1$ on the $j$-th slot. 
In fact, $ \left[\frac{d}{dt} f(\bm{r}_1 + t\bm{e}_j) \right]_{i} =  \left[ \pde{f}{\bm{r}} (\bm{r}_1 + t \bm{e}_j) \bm{e}_j \right]_{i} <0 $, which contradicts the monotonicity assumption.
Hence it must be $ \pde{f}{\bm{r}} (\bm{r}) \geq 0$ $ \forall \,  \bm{r} \in \mathbb{R}^n $. 
Since $ S^{-1} =S $, in the original basis this corresponds to $ S \pde{f}{\bm{z}}(\bm{z})  S \geq 0 $, which is an equivalent condition to $ \mathcal{G}\left( \pde{f}{\bm{z}}(\bm{z}) \right)$ being structurally balanced. \\
Viceversa, if $ \mathcal{G}\left( \pde{f}{\bm{z}}(\bm{z}) \right)$ is structurally balanced, then there exists a gauge transformation $S$ such that, in the new basis $ \bm{r} = S \bm{z} $, $ \pde{f}{\bm{r}} (\bm{r}) \geq 0$ $ \forall \,  \bm{r} \in \mathbb{R}^n $. But then according to \eqref{eq:monot-df} $ \frac{d}{dt} f(\bm{r}_1 + t (\bm{r}_2 -\bm{r}_1) ) \geq 0$ whenever $ \bm{r}_2 - \bm{r}_1 \geq 0$, which is the definition of monotonicity w.r.t. the usual orthant order, and implies that $ f$ is $ \mathbb{S}$-monotone.
\qed

The following theorem and lemma are instead used in the  section ``Structural balance and monotonicity for DNNs'' of the Results.

\begin{theorem}
\label{thm:monot-bal}
Consider a  DNN \eqref{eq:DNN2} in which all the activation functions have nonnegative (generalized) derivative. 
The DNN map $ f \, :\, \mathbb{R}^n \to \mathbb{R}^n$ is monotone iff $ \mathcal{G}(A) $ is structurally balanced. 
\end{theorem}
\paragraph{Proof of Theorem~\ref{thm:monot-bal}.}
In the DNN  \eqref{eq:DNN1}, using the chain rule for differentiation, at the $ \ell$-th layer we have
\[
\pde{\bm{z}_\ell}{\bm{z}_{\ell -1 }} = \pde{\sigma(\bm{q}_\ell)}{\bm{q}_\ell} \pde{\bm{q}_\ell}{\bm{z}_{\ell-1}}
\]
where $ \bm{q}_\ell = W^{(\ell)} \bm{z}_{\ell-1} + \bm{b}^{(\ell)} $ is the pre-activation state.
Since by assumption $  \pde{\sigma(\bm{q}_\ell)}{\bm{q}_\ell} = \diag \left( \pde{\sigma(\bm{q}_{\ell, i})}{\bm{q}_{\ell, i}}\right)  \geq 0$ (elementwise inequality) holds  for all $ \bm{q}_\ell$ (when the partial derivative is not uniquely defined, like in 0 for ReLU, it can be replaced by a differential inclusion, but the inequality still holds) and since $  \pde{\bm{q}_\ell }{\bm{z}_{\ell-1}} = W^{(\ell)}$, then it must be $ \sign \left( \pde{\bm{z}_\ell}{\bm{z}_{\ell -1 }} \right) = \sign(W^{(\ell)})$ wherever $ \pde{\bm{z}_\ell}{\bm{z}_{\ell -1 }} $ has a nonvanishing entry. 
Extra zeros in $ \pde{\bm{z}_\ell}{\bm{z}_{\ell -1 }} $ can happen because of $  \pde{\sigma(\bm{q}_\ell)}{\bm{q}_\ell}  $: $ \left[  \pde{\bm{z}_\ell}{\bm{z}_{\ell -1 }}  \right]_{ij} \neq 0 $ implies $ \left[ W^{(\ell)}\right]_{ij} \neq 0 $ but not viceversa.

Assembling the layers as in \eqref{eq:A} and computing the $ n \times n $ Jacobian matrix of the DNN \eqref{eq:DNN2}, call it $ F(\bm{z})= \pde{f}{\bm{z}}({A\bm{z}+\bm{b}}) $, then from the previous argument we get
\[
 \left[ \begin{array}{ccccc} 
0 & 0 & 0 &  \hdots & 0   \\ 
\pde{ \bm{z}_1}{\bm{x}} & 0 & 0 & \hdots & 0  \\ 
0 &\pde{ \bm{z}_2}{\bm{z}_1} & 0 &  \hdots & 0   \\ 
\vdots & \vdots  & \ddots & \ddots & \vdots  \\ 
0 & 0 &  \hdots & \pde{ \bm{y}}{\bm{z}_{h-1}} & 0 \\
\end{array} \right]   =	
\left[ \begin{array}{cccccc} 
0 & 0 & 0 &  \hdots & 0 & 0   \\ 
\pde{ \sigma(\bm{q}_1)}{\bm{q}_1}  W^{(1)} & 0 & 0 & \hdots & 0  & 0 \\ 
  &   \pde{ \sigma(\bm{q}_2)}{\bm{q}_2} W^{(2)} & 0 &  \hdots & 0  &  \\ 
 \vdots  & \vdots  &  \ddots & & & \\
\vdots  & \vdots  & & \pde{ \sigma(\bm{q}_{h-1})}{\bm{q}_{h-1}} W^{(h-1)} & \vdots  \\ 
0 & 0 &  & \hdots & W^{(h)} & 0 \\
\end{array} \right] 
\]
i.e.,
\beq
 F(\bm{z}) = \underbrace{\diag \left( 0, \pde{ \sigma(\bm{q}_1)}{\bm{q}_1}  ,   \pde{ \sigma(\bm{q}_2)}{\bm{q}_2}  , \ldots, \pde{ \sigma(\bm{q}_{h-1})}{\bm{q}_{h-1}}, I \right)}_{\geq 0} A, \qquad \forall \, \bm{z} \in \mathbb{R}^n
\label{eq:Jacob-monot1}
\eeq

Recall from Proposition~\ref{prop:monotone-funct1} that $ f$ is monotone with respect to some orthant order iff $ \mathcal{G}(F(\bm{z}))$ is structurally balanced. Combining with \eqref{eq:Jacob-monot1}, we have that $ f$ is monotone iff $ \mathcal{G}(A)$ is structurally balanced. 

\qed

From Eq.~\eqref{eq:Jacob-monot1} in the proof of Theorem~\ref{thm:monot-bal}, it is worth highlighting the following property of a DNN $ f$, monotone or not.
Denote $ F(\bm{z})= \pde{f}{\bm{z}}({A\bm{z}+\bm{b}}) $ the Jacobian of $ f$ at $ \bm{z}$. 
\begin{lemma}
\label{lemm:Jac}
In a  DNN~\eqref{eq:DNN2} in which all the activation functions have nonnegative (generalized) derivative, it is $ \sign(F(\bm{z}) ) = \sign(\diag(\mathbb{I}(\bm{z})) A) $.
\end{lemma}

The sufficient but not necessary condition linking the structural balance of the graph $ \mathcal{G}(A) $ to the monotonicity of the IO map of the DNN  used in Section~\ref{sec:IO-monot}``Relation with IO monotonicity'' of the Results is formally proven in the following theorem and corollary.

\begin{theorem}
\label{thm:IOmonot-bal}
Consider a  DNN \eqref{eq:DNN2} in which all the activation functions have nonnegative (generalized) derivative. 
If the subgraph $ \mathcal{G}(A)$ is structurally balanced then the IO map $ \bm{y} = \phi(\bm{x}, \bar{\bm{\theta}} )$ is IO monotone.
\end{theorem}

\paragraph{Proof of Theorem \ref{thm:IOmonot-bal}.}
Let $ A^h $ be the $h$-th power of $A$.
If $ \mathcal{G}(A) $ is structurally balanced, then so is $ \mathcal{G}(A^h)$ and with the same gauge transformation. 
In fact, if $ S A^h S \geq 0 $, then $ (SAS)^h= SA^hS\geq 0$, since $ S^{-1}=S$ and $ S^2 =I$.
In absence of residual connections, the gauge transformed adjacency matrix $SAS$ has lower diagonal blocks $  S^{(\ell)} W^{(\ell)} S^{(\ell -1)} \geq 0$, where $ S^{(\ell)} $ is the diagonal block of $S$ associated with the $ \ell$-th layer.
Furthermore, since
\[
A^h = \begin{bmatrix} 0 & 0 & \ldots & 0 \\ \vdots \\
 W^{(h)}  W^{(h-1)} \ldots W^{(2)}   W^{(1)} & 0 & \ldots & 0 \end{bmatrix},
\]
from $ S A^h S \geq 0$ it must be $ S^{(h)} W^{(h)}  W^{(h-1)} \ldots W^{(2)}   W^{(1)} S^{(0)}\geq 0$, where $ S^{(0)} $ and $ S^{(h)}$ are the diagonal subblocks of $S$ associated to input and output layers, $ S^{(0)}=\diag (\bm{s}_{\bm{x}})$,  $ S^{(h)}=\diag (\bm{s}_{\bm{y}})$.

Following arguments similar to those used in the proof of Theorem~\ref{thm:monot-bal}, let us apply the gauge transformation in input and output space. 
Denote $ \bm{r} = S \bm{z}$ and $ \bm{p} = S \bm{q}$.
If $ \bm{u} = S^{(0)} \bm{x} = \bm{r}_0$ and $ \bm{v} = S^{(h)} \bm{y} = \bm{r}_h$, then after the gauge transformation the input-output map becomes $ \bm{v} = \phi_S(\bm{u}, \bar{\bm{\theta}}) $, or extensively, 
\begin{align*}
\bm{v} & = S^{(h)} W^{(h)} S^{(h-1)}  \bm{r}_{h-1}+ S^{(h)}\bm{b}^{(h)} \\
\bm{r}_\ell &= \sigma( \bm{p}_\ell) = \sigma(S^{(\ell)} W^{(\ell)} S^{(\ell-1)} \bm{r}_{\ell-1} + S^{(\ell)} \bm{b}^{(\ell)}), \qquad  \ell=1, \ldots, h-1 .
\end{align*}
At a point $ \bm{u}_1 $ let us consider an increment $ \bm{u}_2 - \bm{u}_1 >0 $ and compute the directional derivative along $ \bm{u}_2 - \bm{u}_1 $:
\beq
\frac{d}{d t} \phi_S(\bm{u}_1 +t (\bm{u}_2 - \bm{u}_1 )), \bar{\bm{\theta}}) = \pde{\phi_S((\bm{u}_1 + t (\bm{u}_2 - \bm{u}_1 )) }{\bm{u}}(\bm{u}_2 - \bm{u}_1 ) 
\label{eq:der-IO-monot}
\eeq
The Jacobian matrix can be expressed as 
\begin{align*}
\pde{\phi_S}{\bm{u}} 
& = 
\pde{\bm{r}_h}{\bm{p}_h }\pde{\bm{p}_h}{\bm{r}_{h-1} }
\pde{\bm{r}_{h-1}}{\bm{p}_{h-1} }\pde{\bm{p}_{h-1}}{\bm{r}_{h-2} }
\ldots
\pde{\bm{p}_2}{\bm{r}_1} \pde{\bm{r}_1}{\bm{p}_1 }\pde{\bm{p}_1}{\bm{u} } \\
& =
S^{(h)} W^{(h)} S^{(h-1)} 
\pde{\sigma(\bm{p}_{h-1})}{\bm{p}_{h-1}}
S^{(h-1)} W^{(h-1)} S^{(h-2)} \ldots S^{(2)} W^{(2)} S^{(1)}\pde{\sigma(\bm{p}_1)}{\bm{p}_1}S^{(1)} W^{(1)} S^{(0)} 
\end{align*}
Since $ \pde{\sigma(\bm{p}_\ell)}{\bm{p}_\ell} = \diag \left( \pde{\sigma(\bm{p}_{\ell,i})}{\bm{p}_{\ell,i}} \right) \geq 0  $ and $ (S^{(\ell)})^2=I $, $ \pde{\phi_S}{\bm{u}} $ has the same sign pattern as $ S^{(h)} W^{(h)}  W^{(h-1)} \ldots W^{(2)}   W^{(1)} S^{(0)} $, which we  have shown above to be nonnegative. 
Hence in \eqref{eq:der-IO-monot}, since $ \bm{u}_2-\bm{u}_1 \geq 0 $, it must be $ \frac{d}{d t} \phi_S(\bm{u}_1 +t (\bm{u}_2 - \bm{u}_1 )), \bar{\bm{\theta}}) \geq 0$, i.e., along any growing direction of the input the output grows. 
The argument implies that the map $ \phi_S $ is IO monotone with input and output orders which are the usual ones in $ \mathbb{R}^{n_0} $ and $ \mathbb{R}^{n_h}$. 
In the original basis, $ \phi$ is therefore IO monotone with orders $ \mathbb{S}_{\bm{x}} $ and $ \mathbb{S}_{\bm{y}}$. 
\qed

The proof holds also when there are multiple branches in the DNN (e.g. in presence of residual connections). With the modifications made to \eqref{eq:DNN1} for this case (see last paragraph of Section~\ref{app:weight-constr} below), more terms will appear in the expression for $\pde{\phi_S}{\bm{u}}$, all of which are non-negative due to $ S A S \geq 0$ and $\pde{\sigma(\bm{p}_\ell)}{\bm{p}_\ell} \geq 0 $.

Combining Theorem~\ref{thm:monot-bal} with Theorem~\ref{thm:IOmonot-bal} we have the following corollary.
\begin{corollary}
\label{cor:IOmonot-monot}
Under the assumptions of Theorem~\ref{thm:IOmonot-bal}, if the DNN map $ f \, :\, \mathbb{R}^n \to \mathbb{R}^n$ is monotone then the IO map $ \bm{y} = \phi(\bm{x}, \bar{\bm{\theta}} )$ is IO monotone. 
\end{corollary}

\section{Heuristic algorithm for computing frustration: pseudocode} 
\label{app:algorithm}

Algorithm~\ref{alg:frustration} provides the pseudocode for the heuristic procedure mentioned in Section~\ref{sec:algorithm}``Heuristic algorithm for computing frustration'' of the Methods. 

Consider the weighted adjacency matrix $A$ and its symmetrized version $ A_u = A+A^\top $. 
In terms of \eqref{eq:compact-energy} of the paper, the algorithm is initialized at $ S = \diag(\mathds{1})$ and follows a basic gradient descent rule in $S$. It works by computing the row/column sums in $ S A_u S $ and by flipping the sign to the most negative such row/column sum.
If this corresponds to the $ i$th row/column, then the sign flipping is accomplished by flipping the sign of the $i$-th entry in the diagonal of $S$.
In this way the sum $ \mathds{1}^\top S A_u S \mathds{1}$ increases monotonically (i.e., $ e(\bm{s})$ in \eqref{eq:compact-energy} decreases monotonically) as long as there are negative rows/columns in $ S A_u S $. 
It stops when $ S A_u S $ no longer has negative rows/columns.
As the algorithm is heuristic, this is a local minimum in the energy landscape functional $ e(\bm{s})$. 
The local search can be repeated from different initializations, corresponding to choosing different initial $ S = \diag(\bm{s})$.
The best among such local minima is our approximation of the frustration of $A$.

\begin{algorithm}[h!]
\caption{Frustration heuristic}\label{alg:frustration}
\begin{algorithmic}
\Require Adjacency matrix $A\in \mathbb{R}^{n \times n}$ of DNN
\Require Number of initial random sign flips, $\nu \geq 0$
\Require Maximum number of iterations, $M$
\State $ A_u \gets A + A^\top $
\State $S \gets$ gauge transf. with $\nu$ negative entries chosen at random
\State rowsum $\gets S A_u S \mathds{1}$

\For{$M$ iterations}
    \If{$\nexists \; k : $ rowsum$[k]<0$}
        \State \textbf{break}
    \EndIf
    
    \State $i \gets \arg\min_k $ rowsum$[k]$ \Comment{Optionally: choose any $i$ s.t. rowsum$[i]<0$}
    \State $ S_{ii} \gets - S_{ii} $
    \State rowsum $\gets S A_u S \mathds{1}$
\EndFor
\State $\alpha \gets 1/ \sum_{i,j} [\lvert A_u \rvert]_{ij} $
\Ensure $ \hat{\epsilon} \gets \frac{1}{2}\left( 1 - \alpha \mathds{1}^T S A_u S \mathds{1} \right)$
\end{algorithmic}
\end{algorithm}

Notice that the algorithm provides a simple way to localize edges and nodes involved in frustrated cycles, as they correspond to the residual negative edges remaining in the best local minimum matrix $SA_uS$. In fact, the optimal spin assignment in such local minimum is simply $ \mathds{1}$. Reconstructing the original optimal spin assignment in the original basis is also straightforward, as it corresponds to the diagonal entries of the matrix $S$ of all ``accumulated'' single-spin gauge transformations.

\section{Pretrained models}\label{app:models}

The pretrained models we use are retrieved from PyTorch Hub and \textsc{Matlab}'s Deep Learning Toolbox.\footnote{https://www.mathworks.com/help/deeplearning/ref/imagepretrainednetwork.html, https://pytorch.org/hub/pytorch\_vision\_alexnet, and https://pytorch.org/hub/pytorch\_vision\_googlenet} From \textsc{Matlab} we get ShuffleNet, SqueezeNet, ResNet18, AlexNet and GoogLeNet, and from PyTorch we get versions of AlexNet and GoogLeNet with distinct weights and slightly different architecture. The version of ResNet18 available from PyTorch is identical to its counterpart in \textsc{Matlab}, and we therefore disregard it.
The networks have learnable parameters in the span of 1-60 million, but when constructing the adjacency matrices for the networks, keeping in mind that the weights are repeatedly used in convolutions, the number of nodes is in the order of millions, and the number of edges is in the order of billions. See Table~\ref{tab:nnsize} for the exact numbers.

\begin{table}[h]
    \centering
    \caption{Sizes of neural networks used in the experiments. Number of nodes in the adjacency matrix $A$, number of edges (entries in $A$), and number of learnable parameters of the neural networks.\\}
    \label{tab:nnsize}
    \begin{tabular}{lccccc}
        \toprule
        Network & n. nodes & n. edges & n. param. &  \underline{ n. edges } & \underline{ \; n. edges \; } \\
        & $(\times 10^{6})$ & $(\times 10^{9})$ & $(\times 10^{6})$ & n. nodes & n. param. \\
        \midrule
        ShuffleNet & 3.91 & 0.28 & 1.4 & 72 & 200.9 \\
        SqueezeNet & 3.27 & 0.38 & 1.2 & 116 & 306.4 \\
        ResNet18 & 3.01 & 2.80 & 11.7 & 928 & 239.3 \\
        AlexNet & 0.94 & 0.67 & 61.0 & 718 & 11.0 \\
        AlexNet (PyTorch) & 0.73 & 0.66 & 61.1 & 899 & 10.8 \\
        GoogLeNet & 4.80 & 1.52 & 7.0 & 316 & 216.7 \\
        GoogLeNet (PyTorch) & 4.80 & 1.45 & 6.6 & 302 & 218.1 \\
        \bottomrule
    \end{tabular}
\end{table}

\section{Construction of the adjacency matrices of the DNNs}
\label{app:weights}

In \eqref{eq:DNN1} a layer is defined as contributing a matrix of weights $ W^{(\ell)}$, a vector of biases $ \bm{b}^{(\ell)}$ and an activation function $ \sigma(\cdot)$. 
In practice, however, it is common in the literature to refer to each operation that appears in the DNN as a ``layer''.
In particular, the DNNs we consider in this study consist of the following ``layers'': (group) convolutional, dense, max-pooling, average pooling, ReLU, softmax, batch normalization, channel normalization, dropout, and also ``layers'' which are ways of joining parallel branches of the network, such as depth concatenation and addition layers.
We now give some detail on how each of them can be mapped into the representation \eqref{eq:DNN1} of the paper. 
The aim is to provide a precise and reproducible way to construct an adjacency matrix for the DNN, accounting for all layers of the DNNs we use in the experiments.  
Notice that this sometimes requires us to overload a bit the notation of \eqref{eq:DNN1}.

\subsection{Construction of the weight matrices}\label{app:weight-constr}

In this section we detail how the weight matrices of dense, convolutional and pooling layers are constructed for the real network. 
In our setting, layers such as ReLU, normalization and softmax do not contribute any weights to the representation \eqref{eq:A}, see Section~\ref{sec:act-func} for how activation functions are treated.
For instance, the normalization layers only scale preactivations with a positive number (and possibly add an offset term), and they do not change the size of the layer input like pooling layers can.
The pooling layers contribute both (non-trainable) weight matrices, discussed in this section, and an operation represented as an activation, discussed in Section~\ref{sec:act-func}.

In dense (i.e., fully connected) layers, the weight matrices are obtained directly from their usual representation as matrix-vector multiplication and no activation function is present, i.e., $ \bm{z}_{\ell } =  \bm{q}_{\ell } = W^{(\ell)} \bm{z}_{\ell-1} + \bm{b}^{(\ell)} $ with $ W^{(\ell)} $ full. 

Convolutional layers can also be represented in this way, with $ W^{(\ell)} $ which are generalized Toeplitz matrices with sparse structure characterized by repeated entries patterned along shifting rows/columns according to the input size and the stride of each convolutional filter.
More in detail, consider a convolutional layer (index $\ell$) and denote its weight matrix $ W^{(\ell)} \in \mathbb{R}^{n_\ell \times n_{\ell-1}} $. 
For this layer, denote the input and output (without activation) tensors  $\bm{t}_{\ell-1}$ and $\bm{t}_{\ell}$, and let the number of elements in these tensors be $ n_{\ell-1} $ and $ n_{\ell} $ respectively. In the adjacency matrix of the entire network each element corresponds to a node in the graph. To obtain $ W^{(\ell)} $ we vectorize the input and output tensors, and put on each row of $ W^{(\ell)} $ the entries of the kernel such that ${\rm vec}(\bm{t}_{\ell}) =  W^{(\ell)} {\rm vec}(\bm{t}_{\ell-1}) $, see the sketch in Fig.~\ref{fig:construct-dag}.

A more general convolutional layer is the so-called grouped convolutional layer, which separates the input into even groups along the channel/depth dimension, and convolves each group with different filters (allowing for parallel computation) whereafter these output groups are concatenated. Constructing the weight matrix for group convolutions is otherwise entirely analogous to a regular convolution. 
One can shuffle the output channels of a group convolutional layer, as in ShuffleNet, to avoid disentangling the channel groups entirely. Such a channel shuffling layer reorders the $g\cdot m$ channels into $g$ groups of $m$ channels, and it outputs $m$ groups of $g$ channels, where the $i$-th channel in output group $j$ is the $j$-th channel of input group $i$ \cite{zhang2018shufflenet}. When constructing the weight matrix of a layer followed by channel shuffling we reorder the rows of $W^{(\ell)}$ to account for this.

Since pooling operations are convolutions of the feature map with a (max or average) pooling window, the adjacency matrix for a pooling layer is constructed analogously to that of a convolution layer. 
Denote $ \mathcal{N}_{\ell,i} $ the pool of node $i$ at the $\ell$-th layer. 
If the pool is a window of size $ q \times q $, then its cardinality is $  |\mathcal{N}_{\ell,i}| = q^2$. 
We choose to associate to a pooling layer a matrix $ W^{(\ell)}$ whose rows contain up to $ |\mathcal{N}_{\ell,i}| $ nonzero fixed (i.e., not trainable) and identical entries in correspondence of the group of nodes which form the pool of node $i$,  and $ \bm{b}^{(\ell)} = 0 $.
 No distinction is made between average and max-pooling when we consider the entire real networks $A$ (which are independent of the input $ \bm{x}$), while their behavior differ when we look at active subnetworks $ A_{\rm act} (\bm{x})$ in response to a specific input $ \bm{x}$, see Section~\ref{app:active-subn} of this SI. For a pooling window of size $p\times p$, the pooling weights are set to $0.01/p$ to be roughly the same order of magnitude as the neural network weights.

If a convolutional or pooling layer uses zero padding, we do not consider the padded elements as part of the input tensor, as they do not contribute extra edges and they do not appear in the adjacency matrix. Since zero padding egdes are not present, in practice some rows of $W^{(\ell)}$ have fewer edges than the kernel size.

\begin{figure}[h!]
    \centering
    \subfigure[]{
    \begin{minipage}{4.5cm}
    \centering
    \includegraphics[trim=6cm 10.5cm 5cm 9cm, clip=true, width=\textwidth]{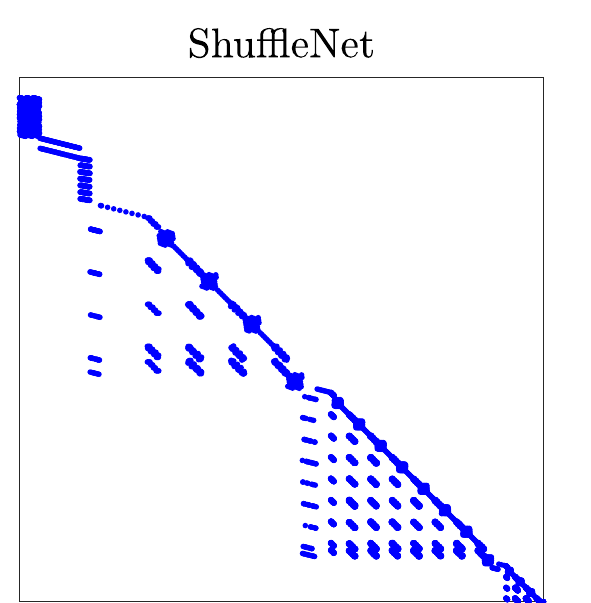}
    \end{minipage}}
    \subfigure[]{
    \begin{minipage}{4.5cm}
    \centering
    \includegraphics[trim=6cm 10.5cm 5cm 9cm, clip=true, width=\textwidth]{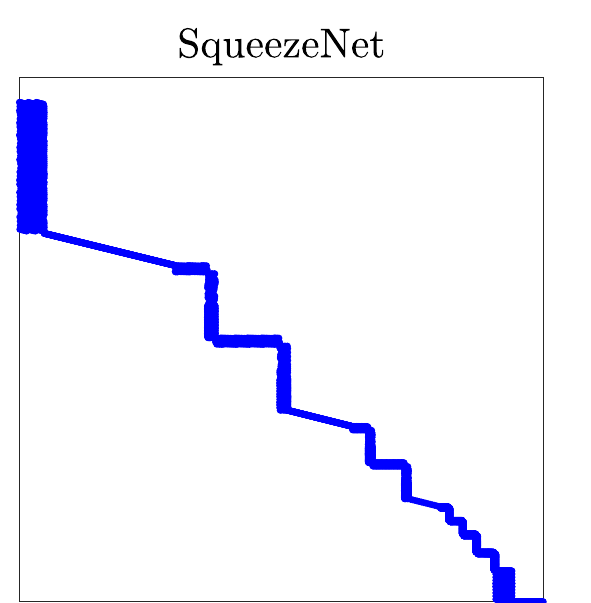}
    \end{minipage}}
    \subfigure[]{
    \begin{minipage}{4.5cm}
    \centering
    \includegraphics[trim=6cm 10.5cm 5cm 9cm, clip=true, width=\textwidth]{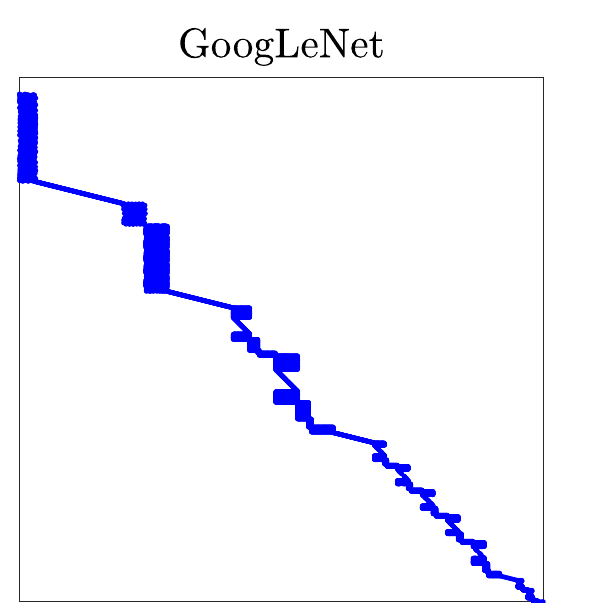}
    \end{minipage}}
    \subfigure[]{
    \begin{minipage}{4.5cm}
    \centering
    \includegraphics[trim=6cm 10.5cm 5cm 9cm, clip=true, width=\textwidth]{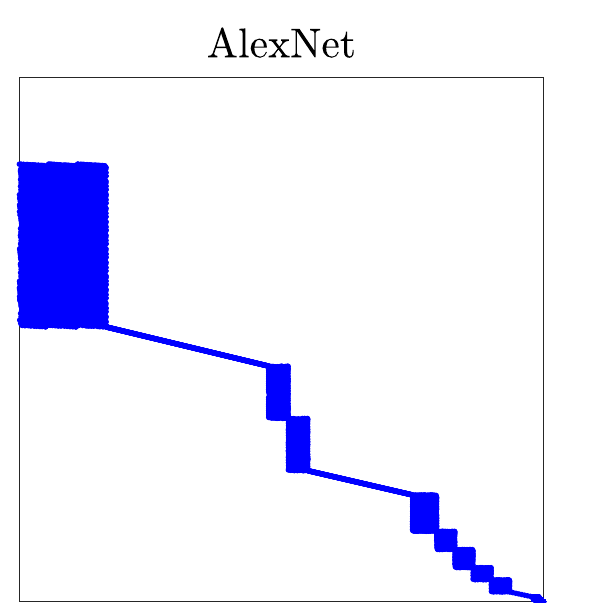}
    \end{minipage}}
    \subfigure[]{
    \begin{minipage}{4.5cm}
    \centering
    \includegraphics[trim=6cm 10.5cm 5cm 9cm, clip=true, width=\textwidth]{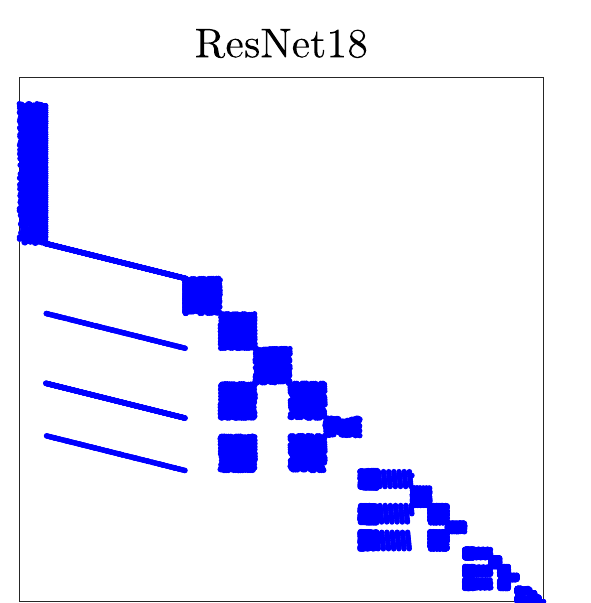}
    \end{minipage}}
    \caption{Adjacency matrix layout for (a) ShuffleNet, (b) SqueezeNet, (c) GoogLeNet, (d) AlexNet, and (e) ResNet18, slightly downsampled for rendering.}
    \label{fig:pretrained_adjmat}
\end{figure}

The weight matrices are then positioned in the adjacency matrix $A$ according to the topology of the neural network. 
When there are no residual connections, then the structure of $A$ is the one shown in \eqref{eq:A} of the paper.
When residual connections are present, i.e., feedforward connections from layers $k_1,...,k_r$ to layer $\ell$, $ k_i< \ell-1 $,  in place of \eqref{eq:DNN1} in the paper, the output of layer $\ell$ is expressed as $ \bm{z}_\ell = \sigma\left( W^{(\ell)} \bm{z}_{\ell-1} + \bm{b}^{(\ell)} + \sum_{i=1}^{r} \bm{z}_{k_i} \right) $. In the adjacency matrix, this means that the weight matrices of layers $k_1,...,k_r$ appear in the block-row corresponding to layer $\ell$, that is, non-zero blocks of weights appear below the lower diagonal blocks in $A$, see visualizations of the adjacency matrices in Fig.~\ref{fig:pretrained_adjmat}, especially items (a) ShuffleNet and (e) ResNet18, where many residual connections are present. Similarly, if the input to layer $\ell$ is a concatenation of the outputs of layers $k_1,...,k_r$, we have $ \bm{z}_\ell = \sigma\left( W^{(\ell)} \begin{bmatrix} \bm{z}_{k_1}^\top & \cdots & \bm{z}_{k_r}^\top \end{bmatrix}^\top  + \bm{b}^{(\ell)}  \right) $, and again the weight matrices appear on the block-row of layer $\ell$, but with each in its own sub-block since now $ W^{(\ell)} \in \mathbb{R}^{n_\ell \times (n_{k_1}+... + n_{k_r})}$. 
In comparison, for addition layer it is $W^{(\ell)} \in \mathbb{R}^{n_\ell \times n_{\ell-1}}$ and $n_{k_1}=...=n_{k_r}=n_{\ell-1}$.

\subsection{A detailed expression for the nonlinearities at the activation functions}
\label{sec:act-func}

Operations occurring in layers such as pooling, ReLU, normalization and softmax can all be interpreted in terms of the activation function $ \sigma(\cdot)$. 

A pooling layer consists of a weight matrix, capturing the neighborhood of a node, and composed as described above, plus a pooling operation which we express as a special case of $ \sigma(\cdot)$.
In max-pooling layers, the max operation is:
\[
z_{\ell,i} = \sigma(\bm{z}_{\ell-1}) =  {\rm max}_{j\in \mathcal{N}_{\ell,i}} z_{\ell-1,j} 
\]
where $ \mathcal{N}_{\ell,i}$ is the pool of node $i$. 
When the pooling operation is averaging, we have instead 
\[
z_{\ell,i} = \sigma(\bm{z}_{\ell-1}) = \frac{1}{| \mathcal{N}_{\ell,i} |} \sum_{j\in \mathcal{N}_{\ell,i}} z_{\ell -1,j} .
\]

Notice that for max-pooling, the overall expression is at odds with the formulation \eqref{eq:DNN1}, in the sense that no matrix-vector multiplication is performed when taking max, and the term $ W^{(\ell)} \bm{z}_{\ell-1} $ should be replaced by a nonlinear dependence $ W^{(\ell)}(\bm{z}_{\ell-1})$.

In a ReLU layer, for each component of $\sigma$ we can represent the activation function as
\[
z_{\ell,i} = \sigma(q_{\ell,i}) = \begin{cases} 
q_{\ell,i}& \text{if  } q_{\ell,i}>0 \\
0  & \text{if  } q_{\ell,i} \leq 0
\end{cases}
\]
where $ q_{\ell,i} $ is the preactivation state of the layer that precedes the ReLU. As a ReLU operation happens on a node, there is no edge weight associated to it.

A batch normalization layer can be represented as 
\[
z_{\ell,i}  = \sigma(q_{\ell,i}) = \frac{\gamma (q_{\ell,i} - \mu_{\mathcal{B}}(\bm{q}_{\ell}))}{\sqrt{\varsigma_{\mathcal{B}}^2(\bm{q}_{\ell}) + \varepsilon}} + \beta
\]
where $\mu_{\mathcal{B}}(\bm{q}_{\ell})$ and $\varsigma_{\mathcal{B}}^2(\bm{q}_{\ell})$ are the mean and variance of $\bm{q}_{\ell}$ in a batch during training (replaced by statistics of the dataset during inference), and $\gamma>0$, and $\beta \in \mathbb{R}$ are parameters learnt during training.

A local response normalization, or cross-channel normalization, was used in \cite{krizhevsky2017imagenet} to promote generalization, and it is part of the \textsc{Matlab} versions of AlexNet and GoogLeNet architectures that we consider in this paper. The layer normalizes the input locally along the channel (third) dimension and is represented by
\begin{equation*}
    z_{\ell,i} = \sigma(\bm{z}_{\ell-1})_{i} = \frac{z_{\ell-1,i}}{K_{\ell-1,i}^\beta}, \qquad K_{\ell-1,i} =\left(K+\frac{\alpha}{w} \sum\limits_{j\in \mathcal{N}_{\ell,i}} z_{\ell-1,j} ^2\right),
\end{equation*}
where $\alpha$, $\beta$, $K$ are positive parameters, $\mathcal{N}_{\ell,i}$ are the node indices in the local normalization window around and including $i$, and $w=\lvert \mathcal{N}_{\ell,i} \rvert$ is the normalization window size. It should be noted that
\begin{align*}
    \frac{\partial \sigma(\bm{z}_{\ell-1})_{i}}{z_{\ell-1,i}} & = \frac{1}{K_{\ell-1,i}^\beta} \left( 1 - \frac{2\alpha \beta}{w K_{\ell-1,i}} z_{\ell-1,i}^2 \right) \\
    \frac{\partial \sigma(\bm{z}_{\ell-1})_{i}}{z_{\ell-1,j}} & = - \frac{2\alpha \beta }{w K_{\ell-1,i}^{\beta+1}}z_{\ell-1,i}z_{\ell-1,j}, \quad j \in \mathcal{N}_{\ell,i} \setminus \{i\}
\end{align*}
which can be negative quantities. However, in practice, the elements of $\bm{z}_{\ell-1}$ (which have been passed through ReLU activations) have magnitudes in the order $10^0 \sim 10^2$, and with the parameters $\alpha = 10^{-4}$, $\beta = 0.75$, $w=5$, $K=1$, it usually means $ \frac{2\alpha \beta z_{\ell-1,i}^2}{wK_{\ell-1,i}}, \frac{2\alpha \beta z_{\ell-1,i} z_{\ell-1,j}}{wK_{\ell-1,i}} \ll 1 $, i.e., that $1 \approx \frac{\partial \sigma(\bm{z}_{\ell-1})_{i}}{\partial z_{\ell-1,i}} > 0 $, and $ \frac{\partial \sigma(\bm{z}_{\ell-1})_{i}}{\partial z_{\ell-1,i}} \gg \frac{\partial \sigma(\bm{z}_{\ell-1})_{i}}{\partial z_{\ell-1,j}} \approx 0$. Hence, $\frac{\partial \sigma(\bm{z}_{\ell-1})}{\partial \bm{z}_{\ell-1}} \approx \diag\left( \frac{\partial \sigma(\bm{z}_{\ell-1})_{i}}{\partial z_{\ell-1,i}}\right) \geq 0$ is a reasonable approximation.

When training a DNN on classification tasks, also the final layer is endowed with an activation function which is usually the softmax function. In Eq.~\eqref{eq:DNN1} and~\eqref{eq:DNN2} of the paper and in the computation of $A$ this layer is ignored, but it is needed when computing the activation adjacency submatrix $ A_{\rm act} (\bm{x})$ of a specific input. 
In terms of $ \sigma$, the softmax function is
\[
\bm{y} = \sigma( \bm{q}_{h} ) = \frac{ e^{\bm{q}_h} }{ \sum_{k=1}^{n_h} e^{q_{h,k} } }
\]
which normalizes the output to $\bm{1}^\top\bm{y}=1$, $\bm{y}\geq 0$, so that it can be interpreted as class probabilities.

In the paper, all these types of operations are indistinctively denoted with the activation function symbol $ \sigma$.

\subsection{Some details for the construction of the activation adjacency submatrix}
\label{app:active-subn}

To construct the adjacency matrix of the active subnetwork induced by a specific input, we give the DNN an input image $ \bm{x}$ and record the hidden states $ \bm{z}$. Starting from the real network, we drop all nodes that are set to zero by ReLU activations and also all edges adjacent to these nodes. All edges that are not active in max-pooling operations are also dropped, i.e., for max-pooling at node $i$ in layer $\ell$ we have $z_{\ell,i} = \max_{j\in \mathcal{N}_{\ell,i}} z_{\ell-1,j}$, and we retain only the edges that connect node $i$ in layer $\ell$ to nodes $k\in \arg\max_{j\in \mathcal{N}_{\ell,i}} z_{<\ell,j} $ in layers connecting to layer $\ell$.
Removing edges between a max-pooling layer and its input layers may lead to nodes in previous layers that are disconnected from the output. We remove such dead-end nodes (and adjacent edges) from the network. For the final layer, we keep only the one node corresponding to the predicted class.

In the paper, the activation pattern in response to an image $ \bm{x}$ is indicated as $ \mathbb{I}(\bm{z})$. 
In practice, this is a bit of an oversimplification.
To be more precise, the pattern $  \mathbb{I} (\cdot) $ depends on the layer type.
For convolutional, dense, and average pooling layers (and no residual connections) we get
\[
W^{(\ell)}_{{\rm act}}  (\bm{x})= {\rm diag}( \mathbb{I}(W^{(\ell)} \bm{z}_{\ell-1}+\bm{b}^{(\ell)}))\, W^{(\ell)} \, {\rm diag}( \mathbb{I}(W^{(\ell-1)} \bm{z}_{\ell - 2} + \bm{b}^{(\ell-1)}))  .
\]
In presence of residual connections, the argument of $ \mathbb{I}(\cdot)$ must be modified accordingly.
As mentioned above, for max-pooling layers we retain only the edges between the output node $i$ and the input node(s) with the maximum value in the pooling window $\mathcal{N}_{\ell,i}$, i.e., the edge(s) which pass on the state. 
Recall that on the full network specifying this properly requires to replace a linear matrix-vector product like $ W^{(\ell)}\bm{z}_{\ell-1} $ with a nonlinear dependence $ W^{(\ell)}(\bm{z}_{\ell-1}) $. 
In the same spirit, the activation submatrix of max-pooling layers can be written as
\[
 [W^{(\ell)}_{{\rm act}}(\bm{z}_{\ell})]_{ij} = \begin{cases}
        [W^{(\ell)}]_{ij} & \text{if } j = \arg \max_{k\in \mathcal{N}_i} z_{\ell,k} \\
        0, & \text{otherwise}
    \end{cases} 
\]
If ReLU activations are present after the max-pooling, the non-activated nodes are dropped just as for other weight layers. 

All networks we consider in this paper have softmax activation after the final layer. When constructing the final layer of the active subnetwork we keep only the edges connected to the node corresponding to the largest output, i.e., the predicted class.
In all cases of interest, $ \mathcal{G}(A_{\rm act}(\bm{x}))$ contains connected components reaching the largest output in $ \bm{y} $ from $ \bm{x}$.

\section{Data availability and acknowledgments}
\label{app:data}

The pretrained CNNs used in this study are all publicly available (with a licence, for the \textsc{Matlab} CNNs), as specified in Section~\ref{app:models} of these SI.
The Imagenette dataset was downloaded from {\tt https://github.com/fastai/imagenette}.
The code we developed can be made available upon request. 

The computations done in the paper were in part enabled by resources provided by the National Academic Infrastructure for Supercomputing in Sweden (NAISS), partially funded by the Swedish Research Council through grant agreement no. 2022-06725.

\begin{figure}
    \centering
    \includegraphics[trim=0cm 9.4cm 0cm 9.3cm, clip=true, width=\linewidth]{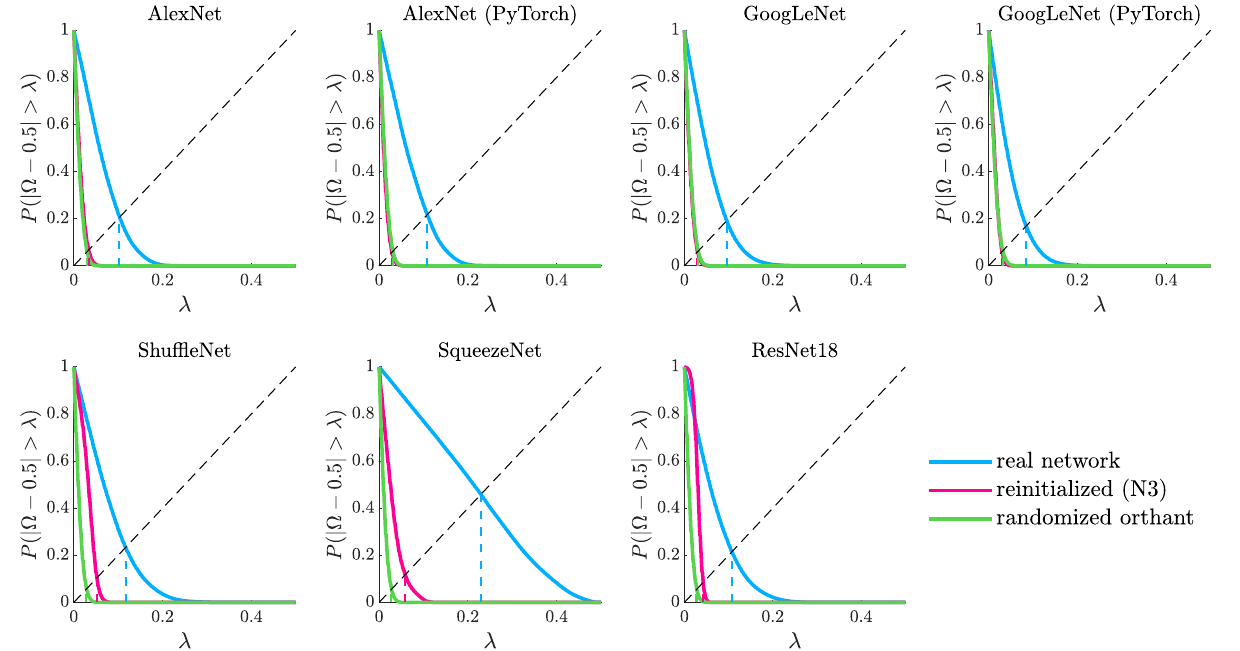}
    \caption{Quantification of the $ \lambda $ coefficient in \eqref{eq:Iomonot2}, for the seven CNNs. In order to compute the output alignment fraction $\Omega$, for each network we consider 100 perturbations for each of 1000 different images. For any value of $\lambda$, the colored line shows the cumulative count of the fraction of cases in which $\Omega$ deviates from 0.5 by more than $\lambda$, i.e., $P(\lvert \Omega - 0.5\rvert > \lambda)$. The IO-function of the network is $\lambda$-monotone if $P(\lvert \Omega - 0.5\rvert > \lambda) \geq 2\lambda$, where the $ 2\lambda $ line is the dashed black line in the figure. The intersection of this line with the cumulative curve gives the value of $ \lambda $ we seek.}
    \label{fig:lambda-panel}
\end{figure}

\begin{figure}
    \centering
    \includegraphics[trim=0cm 9.4cm 0cm 9.3cm, clip=true, width=\linewidth]{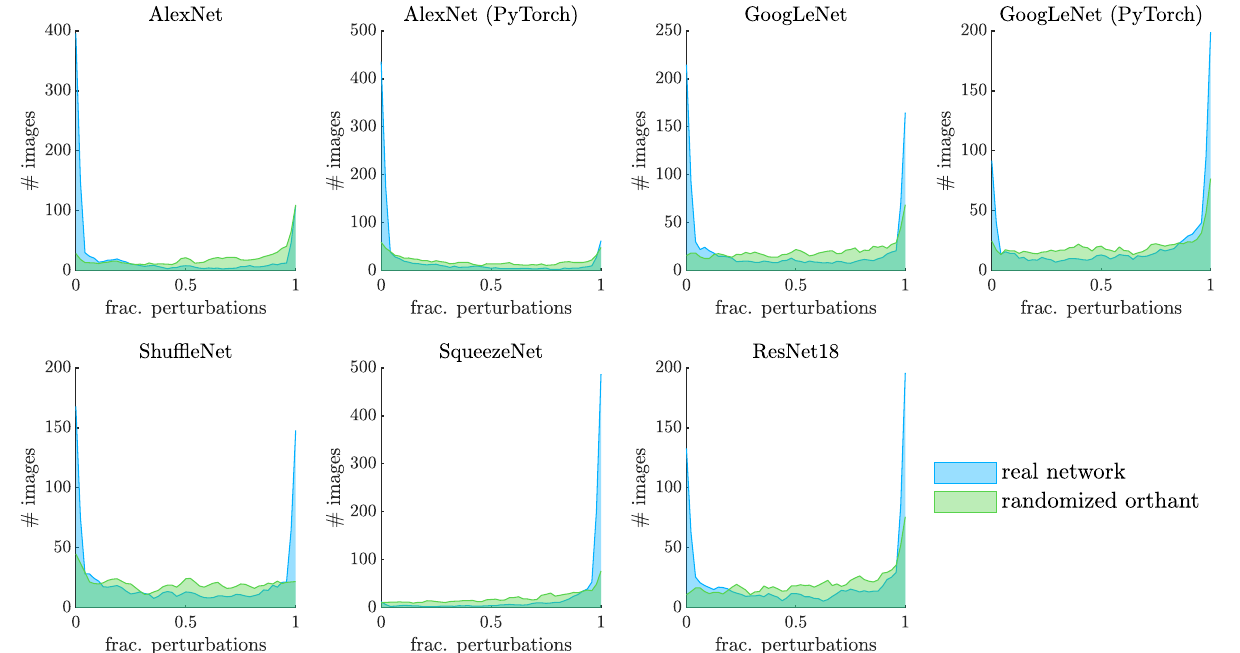}
    \caption{Fraction of perturbations fulfilling $\bm{y}_2 \geq_{\mathbb{S}_y} \bm{y}_1$ in more than 500 (of the 1000) output nodes. In total, 1000 images were used, and 100 perturbations were computed for each image. Values at 0 and 1 mean that all computed perturbations for that image resulted in outputs $\bm{y}_2 \leq_{\mathbb{S}_y} \bm{y}_1$, resp. $\bm{y}_2 \geq_{\mathbb{S}_y} \bm{y}_1$ in more than 500 elements. This corresponds to $\Omega < 0.5$ resp. $\Omega > 0.5$ for all perturbations in Fig.~\ref{fig:Omega-scatter-panel}. For the real networks, the values are concentrated around 0 or 1, indicating that the number of elements increasing (decreasing) in $\mathbb{S}_y$ are always in majority, and do not switch to a majority decreasing (increasing). No such pattern is clearly visible when using random perturbation directions.}
    \label{fig:frac-above05}
\end{figure}

\end{document}